\documentclass{article}

\usepackage[preprint]{neurips_2026}

\usepackage[utf8]{inputenc} 
\usepackage[T1]{fontenc}    
\usepackage{hyperref}       
\usepackage{url}            
\usepackage{booktabs}       
\usepackage{amsfonts}       
\usepackage{nicefrac}       
\usepackage{microtype}
\usepackage{xcolor}         

\usepackage{caption}
\usepackage{graphicx}
\usepackage{amsmath}
\usepackage{xcolor}
\usepackage{amssymb}
\usepackage{amsthm}
\usepackage{algorithm}
\usepackage{algorithmicx}
\usepackage{algpseudocode}
\usepackage{multirow} 
\usepackage{booktabs}
\usepackage{ulem}
\usepackage{wrapfig}

\newtheorem{assumption}{Assumption}
\title{Encryption-Compatible Clustered Federated Learning via Distributed Expectation-Maximization over Metadata}
\author{
Michael Ben Ali \\
UT3, IRIT, CNRS\\
Toulouse, France\\
\texttt{michael-eddy.ben-ali@irit.fr}
\And
Imen Megdiche\\
INU Champollion, ISIS, IRIT, CNRS\\
Castres, France\\
\texttt{imen.megdiche@irit.fr}
\And
André Péninou\\
UT2J, IRIT, CNRS\\
Toulouse, France\\
\texttt{andre.peninou@irit.fr}
\And
Olivier Teste\\
UT2J IRIT, CNRS\\
Toulouse, France\\
\texttt{olivier.teste@irit.fr}
}

\begin{document}

\maketitle
\begin{abstract}
Clustered Federated Learning (CFL) addresses data heterogeneity in federated settings by grouping clients with similar data distributions to enable effective training. Existing methods face a trade-off between privacy preservation, communication cost, and computational efficiency. We formalize this as the CFL trilemma, according to which improving two of these dimensions comes at the expense of the third. A prominent paradigm relies on metadata (i.e., low-dimensional representations of client datasets shared with the server) to enable communication- and computation-efficient clustering. However, such approaches are not compatible with standard FL privacy-preserving mechanisms. To address this limitation, we propose FLAMECHE, which reformulates metadata-based CFL as a distributed Expectation-Maximization (EM) procedure, restricting server updates to additive operations while preserving efficiency. This design enables compatibility with practical secure FL schemes.  We conducted extensive experiments on multiple datasets under various heterogeneous scenarios. Results show that FLAMECHE improves the effectiveness of client models. It enables encryption-compatible metadata-based clustering, enhancing its positioning within the CFL trilemma. 
\end{abstract}
\section{Introduction}
Federated Learning (FL) enables collaborative training of machine learning models without sharing raw data~\cite{mcmahan2017communication}. However, under non-IID (non-independent and identically distributed) data distributions, training a single global model leads to degraded performance~\cite{ye2023heterogeneous}. Clustered Federated Learning (CFL) was introduced to address this limitation by partitioning clients into groups with similar data distributions, enabling the training of more specialized models~\cite{sattler2020clustered}.

Existing CFL methods can be classified into three families~\cite{belfeki2026systematic, ali2025survey}. \textbf{Server-side} approaches~\cite{sattler2020clustered, duan2021flexible, zeng2025stocfl} cluster clients based on similarities between model updates at each communication round, incurring quadratic computational complexity in the number of clients. \textbf{Client-side} approaches~\cite{ghosh2020efficient, long2023multi, FedCAM} shift the clustering process to the clients, where each client must download and evaluate multiple models at each round to determine matching cluster assignment. This increases communication cost and exacerbates the straggler effect, where slower clients delay each round of communication and increase total training time. \textbf{Metadata-based} approaches~\cite{dennis2021heterogeneity, ppcfl, vahidian2023efficient} cluster clients using compact dataset representations, requiring metadata to be shared in plaintext, which raises privacy concerns~\cite{belfeki2026systematic, ali2025survey}. 

 \noindent
\begin{minipage}[c]{0.6\textwidth}
Taken together, these approaches reveal a trade-off in CFL design. Identifying client groups requires either performing complex computations, repetitively refining assignments, or relying on compact representations, each constraining one of three key dimensions: computation, communication, or privacy. We define the \textbf{CFL trilemma} as the difficulty of jointly optimizing these three dimensions within current CFL paradigms. As illustrated in Figure~\ref{fig:cfl_trilemma}, each CFL family lies along one edge of the triangle, favoring two dimensions while sacrificing the third. This tension arises from the clustering task: the server only observes model updates, which are high-dimensional and vary across communication rounds. Inferring stable client groupings requires repeated comparisons over time, increasing server-side computation, while repeated evaluation of candidate models on the client increases communication. Metadata sidesteps these costs by sharing representations directly, at the expense of privacy.

\end{minipage}\hfill
\begin{minipage}[c]{0.38\textwidth}
    \centering
    \includegraphics[width=1\linewidth]{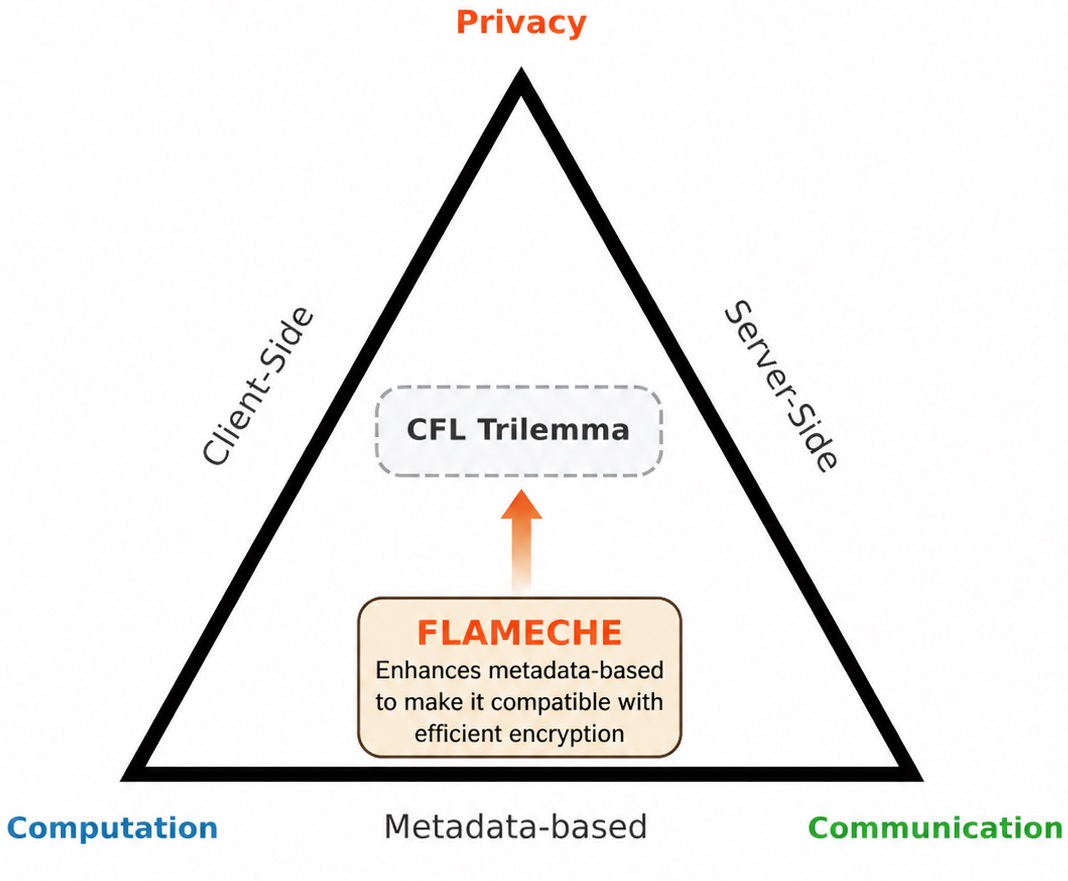}
    \captionof{figure}{\textbf{The CFL Trilemma} is governed by three competing constraints: Privacy, Computation, and Communication.}
    \label{fig:cfl_trilemma}
\end{minipage}

As clustering relies on non-linear operations that are costly under encryption, metadata-based approaches conflict with secure FL mechanisms. These are most efficient under additive-only computations~\cite{bonawitz2017practical, cheon2017homomorphic, zhang2024efficient}, while supporting complex operations inflate computational cost, highlighting the tension of the trilemma. Thus, we investigate the following question. \textbf{Can metadata-based CFL be made more practical under the trilemma issues?}

Our contribution is to reformulate metadata-based clustering as a distributed Expectation-Maximization (EM) procedure, limiting the server to additive operations. Building on this insight, we propose \textbf{FLAMECHE}, a CFL framework designed to cluster clients using metadata while remaining compatible with secure FL.  Metadata is computed using randomly initialized neural networks, avoiding distribution-aware design and prior knowledge of client heterogeneity. 
This solution operates within widely adopted Secure FL constraint (i.e., it neither introduces nor modifies any protocol)~\cite{bonawitz2017practical,wang2024priverifl,cheon2017homomorphic}. It complies with the additive operations efficiently supported by these schemes (e.g., Secure Aggregation and Homomorphic Encryption) while remaining agnostic to their specific implementation.
 The paper is structured as follows. Section~\ref{sec:related_work} gives state-of-the-art insights and issues on CFL. Section~\ref{sec:EMcfl} presents our framework formalization background. Section ~\ref{sec:position} discusses FLAMECHE theoretical costs and its position in the CFL trilemma. Sections ~\ref{sec:results} and~\ref{sec:ablation} challenge our framework against classical CFL methods and strengthen results with an ablation study.
\section{Related Work}~\label{sec:related_work}
\subsection{Clustered Federated Learning for Non-IID Data}
Since the introduction of Federated Learning (FL)~\cite{mcmahan2017communication}, handling non-IID data remains a central challenge~\cite{kairouz2021advances,ye2023heterogeneous,neurocomputingsurvey}. Clustered Federated Learning (CFL) addresses this by grouping clients with similar data distributions and training one model per cluster. Existing methods can be categorized into three families based on how clustering is performed~\cite{belfeki2026systematic, ali2025survey}.

\textbf{Server-side} approaches cluster clients using similarities between model updates. Early works~\cite{ghosh2019robust,briggs2020federated} assume full participation and perform one-shot clustering, while later methods~\cite{sattler2020clustered, duan2021flexible, long2023multi, zeng2025stocfl} extend this to partial participation across rounds. However, updates are high-dimensional, evolve over time, and are computed on different client subsets, requiring repeated clustering. To mitigate this, methods such as FedGroup~\cite{duan2021flexible} use dimensionality reduction (e.g., SVD), FeSEM~\cite{long2023multi} adopts a K-Means-like clustering procedure on model parameters, while StoCFL~\cite{zeng2025stocfl} relies on gradients computed from a frozen anchor model (e.g., the randomly initialized global model) to obtain more stable clustering signals.

\textbf{Client-side} approaches delegate clustering to clients by allowing them to select the most suitable model locally. IFCA~\cite{ghosh2020efficient}, the seminal representative of this family, assigns each client to the model minimizing its local loss. Subsequent client-side methods~\cite{ali2025survey} mainly extend this core mechanism through additional components, such as soft cluster assignments or cross-cluster knowledge transfer, without fundamentally changing the client-driven model selection process. While this avoids explicit server-side clustering, it increases communication by requiring clients to download multiple models each round and raises local computation, potentially leading to client stragglers and higher latency.

\textbf{Metadata-based} approaches rely on compact dataset representations. The effectiveness of these methods is influenced by the choice of metadata. K-Fed~\cite{dennis2021heterogeneity} uses local centroids, while PACFL~\cite{vahidian2023efficient} constructs low-dimensional subspaces. These methods reduce dimensionality and provide stable representations, but typically require sharing metadata with the server, relaxing standard FL privacy assumptions that restrict communication to model updates.


\subsection{Clustering under Privacy Constraints in FL}
The confidentiality of clients' data is fundamental in FL. As, even model updates can reveal sensitive information~\cite{mothukuri2021survey}, several mechanisms have been proposed to strengthen privacy.

\textbf{Differential Privacy (DP)}~\cite{abadi2016deep} perturbs shared information with noise. Its interaction with CFL has received limited attention~\cite{ppcfl, fenoglio2025flux}, and noisy signals may make clustering challenging~\cite{malekmohammadi2025differentially}. Furthermore, DP mechanisms still expose information in plaintext~\cite{mothukuri2021survey}. \textit{Differential privacy in the CFL setting is outside the scope of this work.}

In \textbf{Secure Multi-Party Computation (SMPC)}, multiple parties jointly compute a function over their inputs without revealing them. In FL, protocols such as \textbf{Secure Aggregation}~\cite{bonawitz2017practical} enable practical deployments in which clients apply masks that cancel out during additive aggregation, revealing only the final result. This approach adapts to dynamic FL settings with varying client participation.

\textbf{Homomorphic Encryption (HE)} allows each client to encrypt its data before transmission, enabling the server to perform computations directly on encrypted values without decryption. Schemes such as \textbf{Paillier}~\cite{wang2024priverifl,wang2024fvfl} encryption support only integer additive operations. \textbf{CKKS}~\cite{cheon2017homomorphic,pan2024fedshe} is efficient under linear operations, while non-linear ones incur significant computational overhead.
These constraints have important implications for CFL methods that rely on the server for cluster calculation (i.e., Server-side and Metadata-based CFL). While clustering on encrypted information is theoretically possible, as noted by Zhang et al.~\cite{zhang2024efficient}, ``even clustering low-dimensional representations can lead to prohibitive runtime and memory usage''. Our trilemma highlights that existing approaches favor two dimensions while sacrificing the third. In particular, metadata-based methods achieve top computational and communication efficiency at the cost of relaxed privacy. Improving this privacy dimension remains, to our knowledge, unexplored. 
This gap emphasizes a fundamental limitation: while metadata-based clustering is efficient, it is hardly compatible with encryption mechanisms. This work addresses this limitation by reformulating clustering to restrict on-server operations to additions. 

\section{Efficient Clustered Federated Learning over Encrypted Metadata}~\label{sec:EMcfl}

Consider a federated learning setup with $N$ clients, where each client $i$ holds a local dataset $\mathcal{D}_i = \{(x_{ij}, y_{ij})\}_{j=1}^{n_i}$ drawn from an unknown distribution $P_i(X,Y)$. In CFL, these local distributions are assumed to arise from $K (K < N)$ latent distributions $\{\mathcal{P}_k\}_{k=1}^K$~\cite{sattler2020clustered}. The main challenge in CFL is to cluster clients with similar data, enabling training of specialized
models. To retain the efficiency of metadata-based clustering under privacy-preserving mechanisms, FLAMECHE enforces a key constraint: \textit{all server-side computations are limited to additions}. This ensures compatibility with widely used schemes such as Paillier~\cite{wang2024priverifl}, CKKS~\cite{cheon2017homomorphic}, or Secure Aggregation~\cite{bonawitz2017practical}, where non-linear operations are either unsupported or expensive.

Thus, FLAMECHE decomposes clustering into three steps repeated over communication rounds: (1) the server broadcasts parameters of the $K$ distributions; (2) each client evaluates its likelihood of belonging to each cluster using its metadata; (3) the server aggregates metadata weighted by these likelihoods to update the distribution parameters. This design isolates all non-linear computations to clients while restricting the server to additive operations. This formulation is equivalent to maximum-likelihood estimation in a finite mixture model solved via the Expectation-Maximization (EM) algorithm~\cite{dempster1977maximum}. The E-step is performed locally by clients to compute cluster assignments, while the M-step is executed by the server to update cluster parameters. When the mixture components belong to the exponential family~\cite{dempster1977maximum, murphy2012machine}, the M-step depends only on aggregating sufficient statistics weighted by assignments, and thus requires only additions (Section~\ref{sec:EM}). This preserves the efficiency of metadata-based CFL while enabling compatibility with standard encryption mechanisms.
\subsection{Problem Formulation}

We define a local extractor $\Phi$, which maps a client dataset into a compact vector of dimension $d$:
\[
\Phi : \mathcal{D} \rightarrow \mathbb{R}^d, \quad \phi_i = \Phi(\mathcal{D}_i).
\]
The specific instantiation of the metadata $\phi_i$ is a flexible design parameter (e.g., statistical summaries). 
We model $\{\phi_i\}_{i=1}^N$ as samples drawn from a mixture distribution with $K$ components (each component corresponding to a cluster in CFL), where $\pi_k$ are the mixing coefficients and $\theta_k$ are the cluster-specific parameters. The objective is to estimate the parameters $\Theta = \{\pi_k, \theta_k\}_{k=1}^K$ by maximizing the log-likelihood.
\begin{equation}\label{eq:likelihood}
    \mathcal{L}(\Theta) = \sum_{i=1}^N \log \left( \sum_{k=1}^K \pi_k \, P(\phi_i \mid \theta_k) \right).
\end{equation}

This objective is classically optimized using the Expectation-Maximization (EM) algorithm~\cite{dempster1977maximum}. Thus, we introduce FLAMECHE (Federated Learning Algorithm with Expectation-Maximization Clustering over Hidden Metadata), a distributed EM framework that clusters clients based on their metadata representations $\{\phi_i\}_{i=1}^N$. 
It is not the first FL method to leverage EM, but differs fundamentally in objective and design. FedEM~\cite{dieuleveut2021federated} applies EM at the data level, where latent assignments associate individual samples with a mixture of global models. This objective fundamentally differs from CFM, whose goal is to cluster clients according to their data distributions in order to improve the downstream task. FeSEM~\cite{long2023multi} performs a K-means-like EM procedure over client model parameters, alternating one E-step (client assignment) and one M-step (cluster centroid update) directly on the server at each communication round. In contrast, FLAMECHE performs EM over low-dimensional metadata representations, where the E-step is executed locally on-client while the server performs only the M-step. This design enables compatibility with Secure FL mechanisms.

\subsection{FLAMECHE as a Distributed Expectation-Maximization Algorithm}\label{sec:EM}
To ensure compatibility with additive-only server-side operations, we restrict the clustering model to likelihoods whose M-step can be expressed in terms of additive sufficient statistics. This limits modelization to mixture models from the exponential family :

\begin{assumption}[Exponential Family Mixture Model]
\label{assum:exp_family}
Clients' metadata $(\phi_i)_{1 \leq i \leq N}$ are modeled as samples from a finite mixture of distributions belonging to the exponential family.
\end{assumption}

Under Assumption~\ref{assum:exp_family}, the M-step reduces to computing empirical expectations of sufficient statistics weighted by the responsibilities, which can be expressed as additive aggregations~\cite{dempster1977maximum,murphy2012machine,dieuleveut2021federated}.

Let $t_j(\phi_i)$ denote the components of the sufficient statistics for $j = 1, \ldots, s$, where $s$ is the number of such components. Correspondingly, distributions in the exponential family are fully characterized by their expectation parameters. For instance, in a Gaussian Mixture Model (GMM), this representation includes both the mean and uncentered covariance components, yielding $s = 2$ and $t(\phi_i) = (\phi_i, \phi_i \phi_i^\top)$. Similarly, discrete metadata (e.g. per-class sample counts) can be modeled using a Multinomial mixture. In this case, $\phi_i $ represents the sufficient statistics. 

FLAMECHE clustering proceeds as follows. At the initial step, the server randomly initializes the parameters $\Theta^{(0)}=(\theta_k^{(0)})_{1\leq k \leq K}$. Once each participating client $i$ locally extracts its static metadata vector $\phi_i$, the EM procedure then alternates between two steps at each communication round.
\paragraph{E-Step (Client-Side).}
Given the current global parameter estimates $\Theta^{(r-1)}$ broadcast by the server, at round $r$, each client $i$ computes its responsibilities (i.e., the posterior probabilities that $\phi_i$ belongs to each cluster $k$):
\begin{equation}
    \gamma_{i,k}^{(r)} = \frac{\pi_k^{(r-1)} P(\phi_i \mid \theta_k^{(r-1)})}{\sum_{l=1}^K \pi_l^{(r-1)} P(\phi_i \mid \theta_l^{(r-1)})}, \quad \forall  k \in [1..K].
\label{eq:e_step}
\end{equation}
where $\pi_k^{(r-1)}$ denotes the mixing coefficient  and $\theta_k^{(r-1)}$ the corresponding parameters associated with cluster $k$, both obtained from the M-step at round $r-1$.
For example, in the Gaussian case, $P(\phi_i \mid \theta_k^{(r-1)})$ is computed by evaluating the Gaussian density of cluster $k$ at $\phi_i$.

This step is performed entirely on-client using its metadata and global distribution parameters. While the metadata $\phi_i$ represents a static signature of the local dataset, the responsibilities $\gamma_{i,k}^{(r)}$ update dynamically at each round. Under an encryption mechanism, each client sends to the server $\gamma_{i,k}^{(r)}$ and encrypted vectors $([[ \gamma_{i,k}^{(r)} t_j(\phi_i) ]])_{1\leq j\leq s}$.
\paragraph{M-Step (Server-Side).}
  The server aggregates encrypted vectors using only additive operations:
\begin{equation}
    S_k^{(j)} = \sum_{i=1}^N [[ \gamma_{i,k}^{(r)} t_j(\phi_i) ]], 
    \quad \forall j \in [1..s] \text{,} \quad 
    N_k = \sum_{i=1}^N \gamma_{i,k}^{(r)} \quad \text{and} \quad \pi_k^{(r)} = \frac{N_k}{\sum_{l=1}^K N_l}
\end{equation}

Here, $S_k^{(j)}$ denotes the aggregated sufficient statistics for cluster $k$, and $N_k$ the corresponding effective cluster mass. Each component of the new parameters $\theta_k^{(r)}$ (e.g., means and covariances in case of GMM) are updated via a deterministic mapping of the form $f_j(S^{(j)}_k / N_k)$. For instance, in the Gaussian case, this mapping reduces to computing the mean and covariance as $\mu_k = S^{(1)}_k / N_k$ and $\Sigma_k = S^{(2)}_k / N_k - \mu_k \mu_k^\top$. Importantly, this step does not need to be performed on the server and can be applied after aggregation wherever plaintext values are available (e.g., on the client side at the start of the next E-step). 

While this procedure defines how cluster parameters are estimated, its effectiveness ultimately depends on the quality of the metadata representations $\phi_i$ used for clustering. In particular, the ability to correctly separate client groups relies on how well these representations capture underlying data differences. This raises the following question.
\textbf{How can we design metadata representations to distinguish client groups, without relying on prior knowledge of their data distribution?}

While FLAMECHE offers the flexibility of choosing a metadata, in this paper, we address this challenge with a zero-shot extractor based on randomized neural networks, strictly grounded in the distance-preserving properties of deep networks with random Gaussian weights~\cite{giryes2016deep}.
\subsection{Zero-Shot Metadata Extraction via Randomized Latent Space Projection}\label{sec:gauss}

We construct a compact representation $\phi_i$ of each client dataset $\mathcal{D}_i$ directly from the target model architecture, ensuring alignment with the learning task. For classification, the global model $W$ is decomposed as $W(x)=H(F(x))$, where $F$ is the feature extractor and $H$ the classifier head. We derive metadata from $F$ by truncating the network before the final linear layer, yielding a $d_{\text{feat}}$-dimensional embedding (e.g., $512$ for ResNet-18, $84$ for LeNet-5).

To avoid reliance on trained or pre-trained weights, using a shared random seed, each client initializes $F$ with random Gaussian weights using the well-established Kaiming initialization~\cite{he2015delving}. We justify this zero-shot representation under the following condition.

\begin{assumption}
\label{assum:gaussian_embedding}
Samples from the same distribution exhibit smaller angular separation than samples from different distributions. Under this condition, a randomly initialized ReLU Neural Network (NN)  approximately preserves angular relationships, mapping closer inputs to more similar representations in the latent space.
\end{assumption}

This design is supported by~\cite{giryes2016deep}, which demonstrates that random ReLU networks preserve the angular structure of input. Thus, enabling meaningful representations for grouping without NN training. To summarize each local dataset, we extract metadata $\phi_i$ by computing class-wise empirical means in $F$ induced latent space.
\begin{equation}\label{eq:metadata}
\phi_i = [\mu_{i,1}, \dots, \mu_{i,C}], \text{ with }\mu_{i,c} = \frac{1}{|\mathcal{D}_{i,c}|} \sum_{(x_j, y_j) \in \mathcal{D}_{i,c}} F(x_j), \quad \forall c \in \{1, \dots, C\}
\end{equation}
where $\mathcal{D}_{i,c}$ is the subset of class $c$. This representation has fixed dimension $d = C \cdot d_{\text{feat}}$, independent of dataset size. It is computed once per client, requires no optimization, and remains significantly smaller than model parameters, ensuring low overhead. In case a client has missing labels, it is handled via a simple imputation strategy during E-step (Appendix~\ref{app:missing-label}).

The metadata captures variations in the underlying class-conditional feature distribution $P(X \mid Y)$ rather than differences in label distributions. While we focus on class-wise means, richer statistics (e.g., higher-order moments) could be incorporated. While FLAMECHE is not the first to leverage statistics computed in a feature space induced by a neural extractor~\cite{fenoglio2025flux,tun2023contrastive}, prior approaches rely on stronger assumptions. CP-CFL~\cite{tun2023contrastive} assumes access to a pretrained encoder aligned with the learning task, while FLUX~\cite{fenoglio2025flux} trains the model and requires one full-participation round to align client representations. In contrast, FLAMECHE adopts a fully agnostic approach, relying on randomly initialized extractors and operating strictly under partial participation. Its effectiveness is validated in Section~\ref{sec:ablation}.

\subsection{From EM to a Practical CFL Instantiation}
While the EM framework in Section~\ref{sec:EM} is general, we adopt a practical and efficient instantiation. We consider a spherical K-means-like variant of EM~\cite{dhillon2001concept}, corresponding to a limit case of GMM~\cite{murphy2012machine}. 

The client-side responsibility calculation reduces to $\gamma_{i,k}^{(r)} = \mathbf{1}_{\{k = \arg\min_j \, d(\phi_i, \theta_j^{(r-1)})\}}$, where $\mathbf{1}_{(\cdot)}$ is the indicator function and $d$ the cosine dissimilarity. We adopt this dissimilarity because, unlike Euclidean distance, it focuses on angular differences, which are more stable in high-dimensional representation spaces and better align with Assumption \ref{assum:gaussian_embedding}.

\begin{figure}[H]
    \centering
    \includegraphics[width=0.9\textwidth, trim={0 0 3cm 0}, clip]{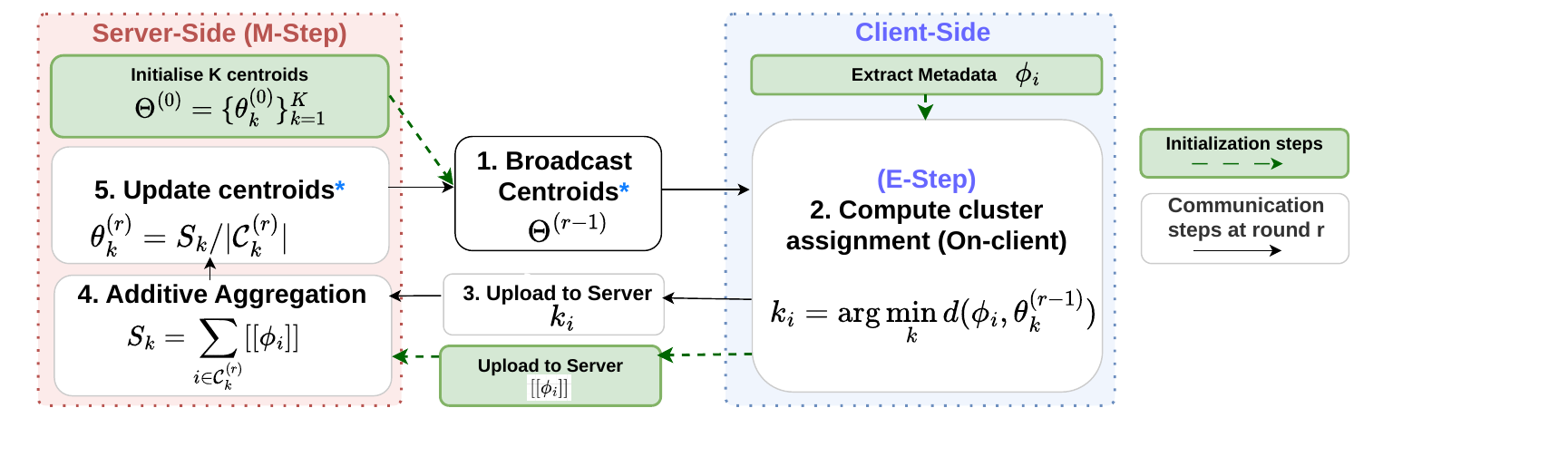}
        \caption{Algorithm~\ref{alg:flameche} clustering mechanism. During initialization, the server initializes centroids and clients extract metadata. At each round $r$: (1) the server broadcasts centroids; (2) clients compute cluster assignments; (3) clients upload metadata (once) and assignments; (4) the server aggregates updates; (5) centroids are updated. \textcolor{blue}{*} Depending on the encryption scheme, the server may broadcast ($S_k$, $|\mathcal{C}^{(r)}_k|$) instead of centroids, allowing clients to compute them locally (see Appendix~\ref{app:crypto}).}
    
    \label{fig:kmeans-flameche}
\end{figure}

For simplified notation, we denote by $k_i$ the latest cluster assignment of client $i$, which is updated whenever the client participates and remains unchanged otherwise, and $\mathcal{C}_{k}^{(r)}$ the set of clients in cluster $k$ at round $r$. Server-side computation reduces to aggregating metadata over cluster members. Each client assigned to cluster $k$ securely contributes its optionally protected metadata $[[\phi_i]]$, and the server computes the aggregated sum.
\begin{equation}\label{eq:M-step}
S_k = \sum_{i \in \mathcal{C}_k^{(r)}} [[\phi_i]], \quad
\theta_k^{(r)} = \frac{S_k}{|\mathcal{C}_k^{(r)}|}, \forall k \in [1..K]
\end{equation}
Depending on the secure mechanism, if $S_k$ is not recovered in plaintext, $\theta_k^{(r)}$ can be computed on-client (see Appendix~\ref{app:crypto}). As metadata remains fixed across rounds, clients upload it only once and subsequently transmit only their cluster assignments in later rounds. We denote as $\mathcal{V}$ the set of seen clients by the server. Each time a client $i$ enters the federation for the first time, the server adds $i$ to $\mathcal{V}$ and the M-step is performed over all clients in $\mathcal{V}$ using their stored metadata and latest assignments.

This practical implementation is detailed in Algorithm~\ref{alg:flameche}. It includes partial participation with rate $p$ (i.e., the clients participating in round $r$ denoted by $\mathcal{P}^{(r)}$). We also introduce a dynamic reclustering strategy to mitigate empty cluster configurations, a well-known issue in EM algorithms~\cite{zhang2003algorithms}. At each round, client $i$ computes the latest error $e_i = d(\phi_i, \theta_{k_i}^{(r-1)})$, corresponding to the distance to its currently assigned centroid. We then perform a periodic check every $\tau$ rounds; the cluster with the highest average error, $k^* = \arg\max_{k \notin E} \frac{1}{|\mathcal{C}_k^{(r)}|} \sum_{i \in \mathcal{C}_k^{(r)}} e_i$, redistribute its clients uniformly among $\mathcal{C}_{k^*}^{(r)}$ and the empty clusters (We denote by  $E$ the set of indices of all empty clusters), followed by an update of the corresponding centroids. For stronger privacy, this procedure can be implemented under the same encryption mechanisms as the metadata. In such a setting, the server is restricted to additive aggregation, while clients assist with decryption and the $\arg\max$ computation. An ablation study of this reclustering strategy is provided in Appendix~\ref{app:recluster}. 

\begin{algorithm}[htb!]
\footnotesize
\caption{FLAMECHE: Federated Learning Algorithm with distributed Expectation-Maximization Clustering over Hidden Metadata}
\label{alg:flameche}
\begin{algorithmic}[1]

\State \textbf{Input:} Number of clusters $K$, Number of rounds $R$, Reclustering frequency $\tau$.

\State

\State Initializes centroids $\{\theta_k^{(0)}\}_{k=1}^K$, models $\{w_k^{(0)}\}_{k=1}^K$, and seen client set $\mathcal{V} = \emptyset$ \Comment{Server}

\For{round $r = 1$ to $R$}

    \State Sample participating clients $\mathcal{P}^{(r)} \subseteq \{1..N\}$ with rate $p$\Comment{Server}

    \State Server broadcasts $\{\theta_k^{(r-1)}\}_{k=1}^K$ to all clients $i \in \mathcal{P}^{(r)}$ \Comment{Server}

    \ForAll{clients $i \in \mathcal{P}^{(r)}$ \textbf{in parallel}}

        \If{$i \notin \mathcal{V}$}

            \State Extract local metadata $\phi_i$ using Equation~\ref{eq:metadata} \Comment{Client}

            \State Upload encrypted metadata $[[\phi_i]]$ to the server once \Comment{Client}

        \EndIf

        \State Compute assignment $k_i = \arg\min_k \, d(\phi_i, \theta_k^{(r-1)})$ \Comment{(\textbf{E-Step}) Client}
        
        \State Compute local assignment error $e_i = d(\phi_i, \theta_{k_i}^{(r-1)})$ \Comment{Client}
        \State Download model $w_{k_i}^{(r-1)}$ and train locally on $\mathcal{D}_i$ to obtain $w_i^{(r)}$ \Comment{Client}
        
        \State Upload model update $w_i^{(r)}$, assignment $k_i$, and error $e_i$ to the server \Comment{Client}

    \EndFor

    \State Update seen clients $\mathcal{V} \leftarrow \mathcal{V} \cup \mathcal{P}^{(r)}$ \Comment{Server} 

    \State Update cluster assignments $\mathcal{C}_k^{(r)} = \{i \in \mathcal{V} : k_i = k\}, \forall k \in [1..K]$  \Comment{Server}

    \For{each non-empty cluster $k$}
        \State Aggregate cluster models 
        $w_k^{(r)} = \frac{1}{|\mathcal{P}^{(r)}_k|} \sum_{i \in \mathcal{P}^{(r)}_k} w_i^{(r)}$ 
        where $\mathcal{P}^{(r)}_k = \mathcal{P}^{(r)} \cap \mathcal{C}_k^{(r)}$ \Comment{Server}

        \State Compute metadata centroid $\theta_k^{(r)}$ using Equation~\ref{eq:M-step} \Comment{(\textbf{M-Step}) Server/Client} 

    \EndFor

    \If{$r \mod \tau = 0$} \Comment{\textbf{(Reclustering Step)} Server}

        \State Set of empty clusters $E = \{k \in [1..K] : \mathcal{C}_k^{(r)} = \emptyset \}$ \Comment{Server}

        \If{$E \neq \emptyset$}

            \State Find cluster with the highest average error 
            $k^* = \arg\max_{k \notin E} \frac{1}{|\mathcal{C}_k^{(r)}|} \sum_{i \in \mathcal{C}_k^{(r)}} e_i$ \Comment{Server/Client}

            \State Distribute the clients of $\mathcal{C}_{k^*}^{(r)}$ uniformly among clusters indexed by $E \cup \{k^*\}$\Comment{Server} 
            \State Recompute $\theta_k^{(r)},$ $\forall k \in E \cup \{k^*\}$ using Equation~\ref{eq:M-step} \Comment{Server} 
        \EndIf
    \EndIf

\EndFor

\end{algorithmic}
\end{algorithm}

\section{Positioning FLAMECHE within the CFL Trilemma}~\label{sec:position}

We position FLAMECHE along the three axes of the CFL trilemma: computational overhead, communication cost, and compatibility with cryptographic constraints.  Table~\ref{tab:complexity_trilemma_overhead_total} reports the additional clustering overhead compared to standard FedAvg~\cite{mcmahan2017communication} of CFL paradigms. Overhead excludes specific secure protocol costs, as approaches are not dependent on them (additional details in Appendix~\ref{app:crypto}). The complexity of Algorithm~\ref{alg:flameche} depends on the metadata dimension $d$ and the number of clusters $K$. Server-side and metadata-based don't take account of clustering cost, as it is algorithm-dependent.  
\begin{table}[h]
\centering
\caption{Added clustering complexity compared to FedAvg over $R$ rounds. $N$ Number of clients, $N_p = p\times N$: Number of participating clients per round, $K$: Number of clusters, $M$: model dimension, $d$: metadata dimension with $d<<M$.}
\label{tab:complexity_trilemma_overhead_total}
\resizebox{0.9\textwidth}{!}{
\begin{tabular}{l cccc}
\toprule
\textbf{CFL Paradigm} & \textbf{Server Computation} & \textbf{Client Computation} & \textbf{Upstream Communication} & \textbf{Downstream Communication} \\
\midrule
Server-side CFL & $\mathcal{O}(R \cdot N_p^2 \cdot M)$ & $-$ & $-$ & $-$ \\
Client-side CFL & $-$ & $\mathcal{O}(R \cdot K \cdot M)$ & $-$ & $\mathcal{O}(R \cdot K \cdot M)$ \\
Metadata-based CFL & $\mathcal{O}(N^2 \cdot d)$ & $-$ & $\mathcal{O}(d)$ & $-$ \\
\midrule
\textbf{FLAMECHE (Ours)} & $\mathcal{O}(R \cdot N_p \cdot K \cdot d)$ & $\mathcal{O}(R \cdot K \cdot d)$ & $\mathcal{O}(d)$ & $\mathcal{O}(R \cdot K \cdot d)$ \\
\bottomrule
\end{tabular}
}
\end{table}

\textbf{Server-side} approaches~\cite{zeng2025stocfl,duan2021flexible} rely on pairwise model similarities, yielding at least $\mathcal{O}(R \cdot N_p^2 \cdot M)$ complexity and requiring non-linear operations incompatible with efficient FL encryption. In contrast, FLAMECHE operates in low-dimensional metadata space with linear scaling $\mathcal{O}(R \cdot N_p \cdot K \cdot d)$, removing both the quadratic dependence on $N_p$ and the reliance on $M$. FLAMECHE’s client-side and communication overhead remains marginal compared to standard model training and transmission.

\textbf{Client-side} methods, represented by IFCA~\cite{ghosh2020efficient}, require evaluating $K$ full models per round, leading to $\mathcal{O}(R \cdot K \cdot M)$ computation and $\mathcal{O}(R \cdot K \cdot M)$ communication. Subsequent client-side variants~\cite{ali2025survey} retain this core mechanism while introducing additional components (e.g., soft assignments or cross-cluster knowledge transfer). In contrast, FLAMECHE performs assignments in metadata space, reducing costs to $\mathcal{O}(R \cdot K \cdot d)$ computation and $\mathcal{O}(R \cdot K \cdot d)$ communication, with $d \ll M$.

\textbf{Metadata-based} approaches~\cite{dennis2021heterogeneity, ppcfl, vahidian2023efficient} achieve low overhead but rely on non-linear server-side clustering, limiting compatibility with standard encryption schemes. FLAMECHE introduces minimal additional overhead while remaining compatible with such mechanisms. Unlike methods that require collecting sufficient metadata before clustering~\cite{dennis2021heterogeneity, ppcfl}, FLAMECHE operates continuously as clients join. The additional server-side overhead is limited to simple additive aggregation.

\textbf{Decoupling Clustering from Training}. FLAMECHE clusters solely rely on static metadata $\{\phi_i\}_{i=1}^N$, independently of model training. This enables exploring clustering configurations (e.g., $K$, metadata design) before training using suitable heuristics (see Appendix.~\ref{sec:tuning}), avoiding repeated FL runs and reducing exploration cost. Overall, FLAMECHE achieves a favorable trade-off in the CFL trilemma by combining low overhead with compatibility with encrypted computation.
\section{Results and Discussion}\label{sec:results}

We evaluate FLAMECHE on five datasets: MNIST, Fashion-MNIST, CIFAR-10 (well-established benchmark in CFL), as well as two real-world medical datasets, TissueMNIST and PathMNIST~\cite{yang2023medmnist}. We simulate three non-IID settings across five random seeds (100 clients, 50 samples per label, 4 latent clusters). These include concept shift on labels, feature distribution skew, and concept shift on features with label skew (a more challenging combined setting). While isolated concept shift via image rotation is generally considered a "solved" baseline in CFL~\cite{zeng2025stocfl,ghosh2020efficient,vahidian2023efficient}, our combined setting deliberately stresses the setup. To induce concept shift in the rotation-invariant medical datasets, we apply channel permutations (PathMNIST) and zooming with grayscale inversion (TissueMNIST). Comprehensive dataset and hyperparameter details are provided in Appendix~\ref{app:result}.
\begin{table}[htb!]
\centering
\caption{Aggregated model performance (accuracy \%) across all non-IID settings. Values represent the pooled mean and standard deviation across the three distinct heterogeneity setups and all random seeds. (s) denote server-side, (c) client-side, and (m) metadata-based methods.}
\label{tab:aggregated_performance}
\resizebox{\textwidth}{!}{ 
\tiny
\begin{tabular}{l ccccc}
\toprule
Algorithm & MNIST & Fashion-MNIST & CIFAR-10 & TissueMNIST & PathMNIST \\
\midrule
Oracle   & $96.29 \pm 1.41$ & $83.70 \pm 1.70$ & $66.05 \pm 7.65$ & $32.08 \pm 5.92$ & $49.09 \pm 7.52$ \\
\midrule
FedAvg   & $81.86 \pm 6.28$ & $66.93 \pm 4.62$ & $54.42 \pm 8.69$ & $21.62 \pm 10.28$ & $30.96 \pm 7.82$ \\
\midrule
FedGroup (s) & $89.52 \pm 10.01$ & $78.30 \pm 10.56$ & $\underline{56.82 \pm 10.39}$ & $29.56 \pm 4.88$ & $44.62 \pm 7.83$ \\
StoCFL (s)   & $83.83 \pm 13.82$ & $69.83 \pm 14.08$ & $44.34 \pm 11.07$ & $24.86 \pm 7.81$ & $30.84 \pm 7.25$ \\
FeSEM (s)    & $90.81 \pm 8.13$ & $78.33 \pm 9.66$ & $54.57 \pm 8.47$ & $25.27 \pm 7.13$ & $43.24 \pm 8.47$ \\
IFCA (c)     & $\underline{92.96 \pm 2.68}$ & $\underline{78.55 \pm 1.99}$ & $51.21 \pm 7.36$ & $30.33 \pm 9.86$ & $33.23 \pm 4.83$ \\
K-Fed (m)    & $89.43 \pm 3.01$ & $73.22 \pm 2.32$ & $48.06 \pm 9.11$ & $\underline{30.76 \pm 4.70}$ & $\mathbf{46.23 \pm 4.90}$ \\
PACFL (m)    & $80.79 \pm 7.13$ & $66.75 \pm 7.99$ & $40.07 \pm 7.24$ & $23.28 \pm 11.60$ & $34.49 \pm 7.20$ \\
\textbf{FLAMECHE (m)} & $\mathbf{94.82 \pm 2.32}$ & $\mathbf{82.61 \pm 2.46}$ & $\mathbf{64.73 \pm 6.59}$ & $\mathbf{32.74 \pm 4.51}$ & $\underline{45.46 \pm 6.87}$ \\
\bottomrule
\end{tabular}
}
\end{table}

We conducted comparisons of FLAMECHE against FedAvg~\cite{mcmahan2017communication}, an Oracle (known partition), server-side methods (FedGroup~\cite{duan2021flexible}, StoCFL~\cite{zeng2025stocfl}, FeSEM~\cite{long2023multi}), the client-side method IFCA~\cite{ghosh2020efficient}, and metadata-based methods (K-Fed~\cite{dennis2021heterogeneity}, PACFL~\cite{vahidian2023efficient}). The selected baselines are restricted to methods that operate strictly under partial participation, without requiring any full-participation rounds, ensuring a fair comparison under consistent system constraints. All models (LeNet-5 or ResNet-18) are trained over 100 rounds with a participation rate $p=20\%$ per round.

\begin{figure}[htb!]
    \centering
    \includegraphics[width=0.85\textwidth]{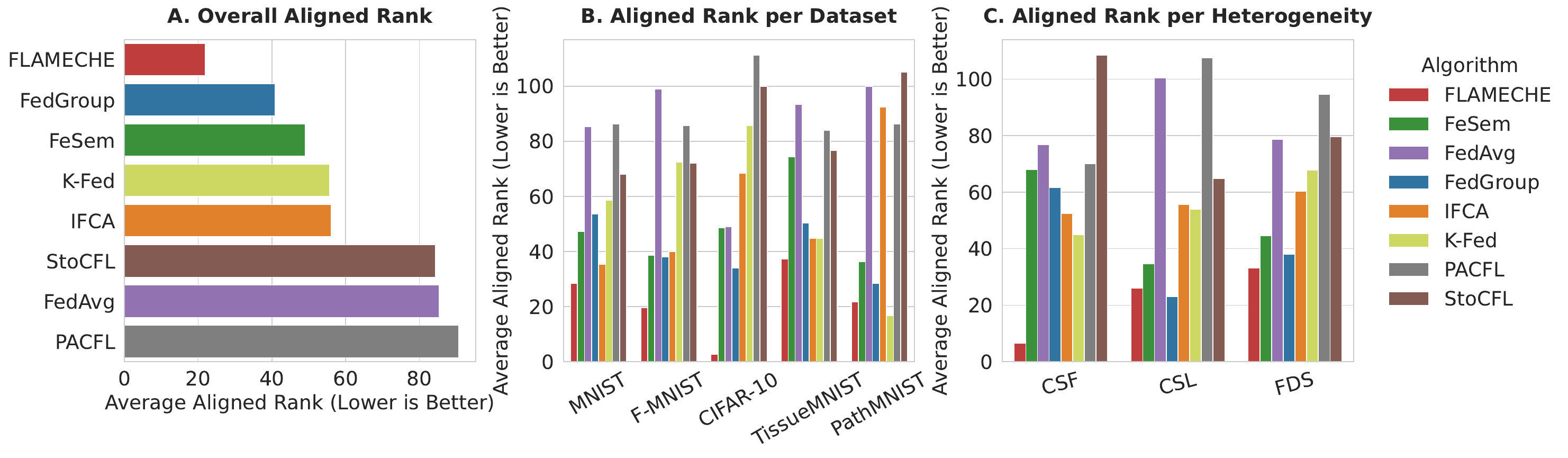}
    \caption{Aligned ranking \cite{liu2022t} across all datasets and heterogeneity settings. FLAMECHE achieves the best overall rank, maintaining consistent performance across concept shift on features (CSF) with label skew, concept shift on labels (CSL), and features distribution skew (FDS). Baselines that perform well on isolated shifts (e.g., PACFL) suffer severe degradation under combined heterogeneity.}
    \label{fig:rank}
\end{figure}

Table~\ref{tab:aggregated_performance} reports the pooled average local test accuracy across all heterogeneity settings, while Figure~\ref{fig:rank} presents aligned rankings~\cite{liu2022t} to emphasize statistically consistent improvements. FLAMECHE achieves the highest overall performance in 4 out of 5 datasets, closely matching the theoretical Oracle. Crucially, FLAMECHE exhibits exceptional robustness (low variance) across different shift types, whereas baselines like FedGroup, StoCFL, and PACFL suffer severe degradation under the combined heterogeneity stress tests. This highlights the fundamental advantage of our approach: conducting EM clustering in a stable, randomized metadata space rather than relying on the noisy, evolving trajectory of model updates.

\section{Ablation Study: Impact of Metadata Representations}\label{sec:ablation}
Table~\ref{tab:metadata_ablation} evaluates FLAMECHE under various metadata representations: our default random neural projections (Gaussian/Uniform), raw class-wise average images similar to K-fed (Mean), reduced averages (Mean + PCA), and subspace representations (PACFL). Clustering quality is measured via ARI (agreement with the Oracle partition) and DBI (cluster compactness, lower is better).

Random projections yield strong stability, and near-optimal clustering $(ARI \approx 0.97)$ across all datasets. While raw Mean images perform similarly in multiple setups, they operate in a higher-dimensional, highly interpretable (and thus privacy-sensitive) space. Conversely, PCA reduction and PACFL-style subspaces fail in this additive EM setting, likely because simple additive aggregation destroys their underlying geometric structures. This confirms that untrained random projections offer a highly effective, zero-shot metadata extraction strategy without requiring dataset-specific engineering.

\begin{table}[htb!] 
\centering
\caption{Impact of FLAMECHE metadata representation on clustering performance. Results are averaged over 5 seeds and 3 heterogeneity settings. }
\label{tab:metadata_ablation}
\resizebox{\textwidth}{!}{ 
\begin{tabular}{l cccccccccc}
\toprule
 & \multicolumn{2}{c}{MNIST} & \multicolumn{2}{c}{Fashion-MNIST} & \multicolumn{2}{c}{CIFAR-10} & \multicolumn{2}{c}{TissueMNIST} & \multicolumn{2}{c}{PathMNIST} \\
\cmidrule(lr){2-3} \cmidrule(lr){4-5} \cmidrule(lr){6-7} \cmidrule(lr){8-9} \cmidrule(lr){10-11}
Metadata & ARI & DBI & ARI & DBI & ARI & DBI & ARI & DBI & ARI & DBI \\
\midrule
\textbf{Gaussian} & $\underline{0.97 \pm 0.04}$ & $\underline{0.60 \pm 0.07}$ & $\mathbf{0.97 \pm 0.04}$ & $\underline{0.47 \pm 0.07}$ & $\mathbf{0.97 \pm 0.04}$ & $\mathbf{1.53 \pm 0.39}$ & $\mathbf{0.97 \pm 0.04}$ & $\mathbf{1.15 \pm 0.53}$ & $\mathbf{0.97 \pm 0.04}$ & $\mathbf{1.04 \pm 0.40}$ \\
Uniform  & $\underline{0.97 \pm 0.04}$ & $\mathbf{0.57 \pm 0.10}$ & $\mathbf{0.97 \pm 0.04}$ & $\mathbf{0.45 \pm 0.08}$ & $\underline{0.94 \pm 0.14}$ & $\underline{1.63 \pm 0.58}$ & $\mathbf{0.97 \pm 0.04}$ & $\mathbf{1.15 \pm 0.51}$ & $\mathbf{0.97 \pm 0.04}$ & $\underline{1.10 \pm 0.43}$ \\
Mean     & $\mathbf{0.98 \pm 0.04}$ & $0.83 \pm 0.09$ & $\mathbf{0.97 \pm 0.04}$ & $0.52 \pm 0.12$ & $\mathbf{0.97 \pm 0.04}$ & $1.91 \pm 0.17$ & $\mathbf{0.97 \pm 0.04}$ & $\underline{1.38 \pm 0.61}$ & $\underline{0.91 \pm 0.13}$ & $1.30 \pm 0.34$ \\
Mean + PCA & $0.21 \pm 0.18$ & $2.53 \pm 0.47$ & $0.57 \pm 0.47$ & $1.65 \pm 0.63$ & $0.43 \pm 0.39$ & $3.04 \pm 0.58$ & $\underline{0.49 \pm 0.32}$ & $2.80 \pm 0.60$ & $0.47 \pm 0.38$ & $2.00 \pm 0.61$ \\
PACFL    & $0.00 \pm 0.01$ & $6.63 \pm 0.64$ & $0.02 \pm 0.04$ & $6.26 \pm 0.48$ & $0.00 \pm 0.02$ & $5.90 \pm 0.95$ & $0.10 \pm 0.16$ & $6.63 \pm 0.64$ & $0.01 \pm 0.02$ & $7.01 \pm 0.20$ \\
\bottomrule
\end{tabular}
}
\end{table}
\section{Conclusion}
By reformulating metadata-based CFL as a distributed EM procedure, FLAMECHE confines non-additive operations to clients, restricting server updates to additive aggregation. This enables compatibility with encryption FL mechanisms without sacrificing efficiency. Evaluations show that our framework delivers accuracy and robustness across diverse and complex data heterogeneities. Even though FLAMECHE provides a practical solution, our analysis is empirical. While our framework is designed to satisfy additive constraints, it does not assume a specific cryptographic infrastructure (see Appendix~\ref{app:crypto}). Although its EM formulation is general, current evaluations are restricted to hard clustering; extending them to soft settings (e.g., GMM) is left for future work. The use of partial participation relaxes standard EM convergence, leaving convergence proof as an open question. Finally, even if metadata can be protected via encryption, it remains a structured summary of local data, and its potential privacy implications should be considered.

\bibliographystyle{plain}
\bibliography{refs}
\appendix

\section{Experimental Details and Additional Results}~\label{app:result}

\paragraph{Experimental Setup.}
We consider $100$ clients, each initially holding balanced local datasets with $50$ samples per label before applying heterogeneity transformations. Clients are partitioned into $4$ groups of $25$, corresponding to $K=4$ latent data-generating distributions.

All experiments are conducted over $100$ communication rounds with a client sampling rate of $20\%$ per round. For each random seed, both the data distribution across clients and the client participation schedule are randomly generated. Each seed therefore jointly determines the client data partitioning and the per-round client sampling sequence. For a given seed, all compared methods share the exact same sampled clients at each round, ensuring strict comparability. Results are reported as the average over $5$ independent random seeds.

We evaluate three heterogeneity settings in the following order: (1) \textit{concept shift on labels}, (2) \textit{features distribution skew}, and (3) \textit{concept shift on features combined with label skew}. While the main paper reports results aggregated across heterogeneity types, we provide here detailed results for each setting (Tables~\ref{tab:concept_labels}, \ref{tab:features_skew}, and \ref{tab:concept_features}).

We use LeNet-5 for MNIST and Fashion-MNIST, and ResNet-18 for CIFAR-10, TissueMNIST, and PathMNIST. Local training is performed for $5$ epochs (LeNet-5) and $10$ epochs (ResNet-18) using the Adam optimizer with learning rate $10^{-3}$ and default parameters. Batch size is set to $128$ for grayscale datasets and $256$ for RGB datasets. All methods use identical architectures and initialization schemes when applicable.

\paragraph{Heterogeneity Construction.}
Heterogeneity is introduced through controlled transformations applied at the group level.

\textit{(1) Concept shift on labels.}
Each group is assigned a specific label permutation. For instance, in CIFAR-10, different groups apply swaps such as $(0 \leftrightarrow 2)$, $(1 \leftrightarrow 7)$, $(0 \leftrightarrow 5)$, and $(4 \leftrightarrow 7)$. This creates distinct label semantics across groups while preserving input distributions.

\textit{(2) Features distribution skew.}
Feature distributions are modified without altering labels. Four groups are constructed using image transformations: erosion with a $3\times3$ kernel, dilation with a $3\times3$ kernel, dilation with an $8\times8$ kernel, and a last group without transformation. This induces distributional skew in the features space while preserving label consistency.

\textit{(3) Concept shift on features + label skew.}
This setting combines feature-level transformations and label imbalance. Feature transformations are dataset-specific: rotations (0°, 90°, 180°, 270°) for MNIST, Fashion-MNIST, and CIFAR-10; morphological and intensity transformations for TissueMNIST (normal, zoomed, inverted grayscale, inverted grayscale with zoom); and channel permutations for PathMNIST. Each concept group is further subdivided into five label distributions: the original (unskewed) distribution, and four skewed variants (normal, anti-normal, left-skewed, and right-skewed). This results in $4 \times 5 = 20$ distinct empirical client distributions overall.
\paragraph{Choice of the Number of Clusters.}
The number of clusters is fixed at $K=4$ for all algorithms that require this parameter as input. This matches the number of underlying class-conditional feature variations ($P(X|Y)$) in our experimental construction. While selecting $K$ is a non-trivial problem in CFL, our goal here is to evaluate clustering methods under controlled and known heterogeneity.

Importantly, in the combined setting (concept shift on features with label skew), the additional label distributions introduce intra-group variability without necessarily corresponding to distinct clusters. Empirically, we observe that modeling $K=4$ clusters remains the most effective choice in this setting (Appendix~\ref{sec:tuning}). We therefore adopt $K$ values as a consistent experimental configuration across all methods, rather than claiming it to be universally optimal.

\paragraph{Evaluation Metrics.}
Algorithm effectiveness is measured by the average test accuracy across clients' test sets. Clustering quality is evaluated using the Adjusted Rand Index (ARI) and the Davies–Bouldin Index (DBI). ARI measures agreement with the ground-truth client partition (used by the Oracle baseline), while DBI evaluates cluster compactness and separation (lower is better). All metrics are reported at the final communication round. 

\paragraph{Baselines.} For all comparisons, we exclusively consider representative \emph{hard clustering} CFL methods~\cite{ali2025survey}, whose primary objective is to discover the underlying client cluster structure. We intentionally exclude approaches combining clustering with additional mechanisms such as personalization, cross-cluster knowledge transfer, or auxiliary optimization modules, as these improvements are orthogonal to the clustering strategy itself and could theoretically be incorporated into most hard clustering methods. Restricting the comparison to pure hard clustering approaches therefore isolates the contribution of the clustering mechanism and enables a fair evaluation.

FedAvg ~\cite{mcmahan2017communication} follows the standard federated averaging procedure. Oracle corresponds to FedAvg trained independently within ground-truth clusters.

\textbf{Server-side methods :} FedGroup ~\cite{duan2021flexible} is implemented using the recommended Euclidean Distance of Cosine dissimilarity (EDC) metric and a cold-start phase using $40\%$ of clients (equivalent to two rounds of participation). StoCFL ~\cite{zeng2025stocfl}) requires threshold selection; we perform a grid search over the full similarity matrix using a binary search procedure to identify the threshold that best separates clients into $4$ clusters, as default values were not suitable in our setting.  FeSEM~\cite{long2023multi} performs a K-means-like clustering procedure on client model parameters.

\textbf{Client-side methods :} IFCA~\cite{ghosh2020efficient} is sensitive to initialization; we run $5$ parallel initializations and retain the model achieving the best validation accuracy. 

\textbf{Metadata-based methods :} K-Fed~\cite{dennis2021heterogeneity} follows a one-shot clustering strategy, in which a global FedAvg model is first trained until metadata (local dataset centroids) from all clients have been collected, then used to perform one-shot clustering. PACFL ~\cite{vahidian2023efficient}) is implemented with the recommended hyperparameters, using $5$ components for subspace decomposition.

\textbf{FLAMECHE: } FLAMECHE uses a randomly initialized feature extractor with Kaiming Gaussian weights, use the same initialization shared across all clients. Reclustering is triggered every $\tau = 10$ rounds.

\paragraph{Results Discussion.}
Detailed results for each heterogeneity setting are reported in Tables~\ref{tab:concept_labels}, \ref{tab:features_skew}, and \ref{tab:concept_features}. While FLAMECHE is not always the top-performing method in every individual configuration, it consistently achieves strong performance across all heterogeneity types.

\begin{table}[htb!]
\centering
\caption{Clustering Performance under Concept Shift on Labels}
\label{tab:concept_labels}
\resizebox{\textwidth}{!}{
\begin{tabular}{l cccccccccc}
\toprule
 & \multicolumn{2}{c}{MNIST} & \multicolumn{2}{c}{Fashion-MNIST} & \multicolumn{2}{c}{CIFAR-10} & \multicolumn{2}{c}{TissueMNIST} & \multicolumn{2}{c}{PathMNIST} \\
\cmidrule(lr){2-3} \cmidrule(lr){4-5} \cmidrule(lr){6-7} \cmidrule(lr){8-9} \cmidrule(lr){10-11}
Algorithm & ARI & Accuracy & ARI & Accuracy & ARI & Accuracy & ARI & Accuracy & ARI & Accuracy \\
\midrule
Oracle   & $1.00 \pm 0.00$ & $97.90 \pm 0.19$ & $1.00 \pm 0.00$ & $85.80 \pm 0.34$ & $1.00 \pm 0.00$ & $75.02 \pm 0.54$ & $1.00 \pm 0.00$ & $37.44 \pm 2.82$ & $1.00 \pm 0.00$ & $49.29 \pm 7.84$ \\
\midrule
FedAvg   & --- & $73.92 \pm 2.95$ & --- & $64.56 \pm 0.62$ & --- & $56.05 \pm 6.92$ & --- & $24.72 \pm 9.93$ & --- & $34.00 \pm 7.64$ \\
\midrule
FedGroup (s) & $\mathbf{1.00 \pm 0.00}$ & $\underline{97.78 \pm 0.19}$ & $\mathbf{1.00 \pm 0.00}$ & $\underline{85.59 \pm 0.33}$ & $0.54 \pm 0.15$ & $\underline{66.07 \pm 3.64}$ & $0.99 \pm 0.02$ & $31.78 \pm 2.93$ & $0.60 \pm 0.15$ & $\mathbf{48.99 \pm 6.72}$ \\
StoCFL (s)& $\mathbf{1.00 \pm 0.00}$ & $96.42 \pm 0.32$ & $\mathbf{1.00 \pm 0.00}$ & $81.38 \pm 0.71$ & $0.33 \pm 0.01$ & $54.89 \pm 3.89$ & $0.80 \pm 0.17$ & $27.90 \pm 0.88$ & $0.62 \pm 0.17$ & $33.84 \pm 7.66$ \\
FeSEM (s)& $\mathbf{1.00 \pm 0.00}$ & $\mathbf{97.90 \pm 0.17}$ & $\mathbf{1.00 \pm 0.00}$ & $\mathbf{85.75 \pm 0.41}$ & $0.31 \pm 0.08$ & $61.25 \pm 2.02$ & $0.78 \pm 0.08$ & $30.76 \pm 2.92$ & $0.45 \pm 0.09$ & $46.46 \pm 3.53$ \\
IFCA (c)& $\mathbf{1.00 \pm 0.00}$ & $95.69 \pm 0.34$ & $\mathbf{1.00 \pm 0.00}$ & $81.31 \pm 0.13$ & $0.55 \pm 0.15$ & $56.89 \pm 1.64$ & $0.34 \pm 0.08$ & $\underline{34.42 \pm 0.53}$ & $0.10 \pm 0.08$ & $35.78 \pm 3.50$ \\
K-Fed (m)& $\mathbf{1.00 \pm 0.00}$ & $93.37 \pm 0.48$ & $\mathbf{1.00 \pm 0.00}$ & $75.83 \pm 1.13$ & $\mathbf{1.00 \pm 0.00}$ & $56.59 \pm 3.02$ & $\mathbf{1.00 \pm 0.00}$ & $33.41 \pm 2.61$ & $\underline{0.67 \pm 0.01}$ & $\underline{47.44 \pm 4.46}$ \\
PACFL (m)& $0.00 \pm 0.01$ & $71.66 \pm 1.34$ & $-0.01 \pm 0.01$ & $55.58 \pm 0.83$ & $0.00 \pm 0.00$ & $43.00 \pm 3.81$ & $0.00 \pm 0.01$ & $25.60 \pm 3.93$ & $0.00 \pm 0.01$ & $34.63 \pm 8.12$ \\
FLAMECHE (m)& $\underline{0.96 \pm 0.05}$ & $95.60 \pm 3.03$ & $\underline{0.97 \pm 0.05}$ & $83.96 \pm 3.22$ & $\underline{0.92 \pm 0.07}$ & $\mathbf{70.94 \pm 4.45}$ & $\mathbf{1.00 \pm 0.00}$ & $\mathbf{34.94 \pm 2.56}$ & $\mathbf{0.90 \pm 0.19}$ & $46.07 \pm 8.68$ \\
\bottomrule
\end{tabular}
}
\end{table}

\begin{table}[htb!]
\centering
\caption{Clustering Performance under Features Distribution Skew}
\label{tab:features_skew}
\resizebox{\textwidth}{!}{
\begin{tabular}{l cccccccccc}
\toprule
 & \multicolumn{2}{c}{MNIST} & \multicolumn{2}{c}{Fashion-MNIST} & \multicolumn{2}{c}{CIFAR-10} & \multicolumn{2}{c}{TissueMNIST} & \multicolumn{2}{c}{PathMNIST} \\
\cmidrule(lr){2-3} \cmidrule(lr){4-5} \cmidrule(lr){6-7} \cmidrule(lr){8-9} \cmidrule(lr){10-11}
Algorithm & ARI & Accuracy & ARI & Accuracy & ARI & Accuracy & ARI & Accuracy & ARI & Accuracy \\
\midrule
Oracle   & $1.00 \pm 0.00$ & $94.53 \pm 0.23$ & $1.00 \pm 0.00$ & $83.39 \pm 0.40$ & $1.00 \pm 0.00$ & $66.61 \pm 1.01$ & $1.00 \pm 0.00$ & $31.43 \pm 2.87$ & $1.00 \pm 0.00$ & $52.40 \pm 3.58$ \\
\midrule
FedAvg   & --- & $88.00 \pm 1.52$ & --- & $72.82 \pm 0.53$ & --- & $\underline{61.56 \pm 2.62}$ & --- & $27.09 \pm 8.35$ & --- & $28.90 \pm 9.98$ \\
\midrule
FedGroup (s)& $\mathbf{1.00 \pm 0.00}$ & $\mathbf{94.43 \pm 0.15}$ & $\mathbf{1.00 \pm 0.00}$ & $\underline{83.37 \pm 0.25}$ & $0.28 \pm 0.07$ & $61.47 \pm 1.59$ & $0.07 \pm 0.05$ & $31.89 \pm 4.47$ & $\underline{0.99 \pm 0.02}$ & $46.38 \pm 7.09$ \\
StoCFL (s) & $0.88 \pm 0.16$ & $90.40 \pm 0.98$ & $\mathbf{1.00 \pm 0.00}$ & $78.07 \pm 0.63$ & $0.08 \pm 0.17$ & $47.09 \pm 6.30$ & $0.04 \pm 0.00$ & $32.25 \pm 0.00$ & $0.71 \pm 0.19$ & $32.31 \pm 7.75$ \\
FeSEM (s)& $0.95 \pm 0.12$ & $\underline{94.39 \pm 0.58}$ & $\underline{0.95 \pm 0.13}$ & $\mathbf{83.41 \pm 0.68}$ & $0.18 \pm 0.10$ & $59.62 \pm 1.41$ & $0.08 \pm 0.09$ & $24.55 \pm 9.48$ & $0.77 \pm 0.14$ & $\mathbf{49.53 \pm 5.85}$ \\
IFCA (c)& $0.64 \pm 0.15$ & $89.58 \pm 1.40$ & $0.76 \pm 0.12$ & $77.26 \pm 0.51$ & $0.48 \pm 0.12$ & $54.53 \pm 2.63$ & $0.19 \pm 0.00$ & $\mathbf{39.83 \pm 0.00}$ & $0.11 \pm 0.04$ & $36.38 \pm 1.40$ \\
K-Fed (m)& $\mathbf{1.00 \pm 0.00}$ & $86.36 \pm 0.28$ & $\mathbf{1.00 \pm 0.00}$ & $72.38 \pm 1.06$ & $\mathbf{1.00 \pm 0.00}$ & $51.68 \pm 0.51$ & $\mathbf{1.00 \pm 0.00}$ & $32.11 \pm 4.05$ & $\mathbf{1.00 \pm 0.00}$ & $\underline{48.49 \pm 4.51}$ \\
PACFL (m)& $0.27 \pm 0.02$ & $81.86 \pm 0.79$ & $0.73 \pm 0.13$ & $72.28 \pm 1.50$ & $0.01 \pm 0.00$ & $43.67 \pm 1.42$ & $0.01 \pm 0.02$ & $33.66 \pm 10.42$ & $0.21 \pm 0.12$ & $31.65 \pm 4.73$ \\
FLAMECHE (m)& $\underline{0.97 \pm 0.04}$ & $93.92 \pm 1.09$ & $\underline{0.95 \pm 0.05}$ & $82.86 \pm 1.04$ & $\underline{0.98 \pm 0.04}$ & $\mathbf{66.19 \pm 1.06}$ & $\mathbf{1.00 \pm 0.00}$ & $\underline{33.73 \pm 6.20}$ & $0.98 \pm 0.04$ & $46.20 \pm 7.35$ \\
\bottomrule
\end{tabular}
}
\end{table}

\begin{table}[h]
\centering
\caption{Clustering Performance under Concept Shift on Features (Label Skew)}
\label{tab:concept_features}
\resizebox{\textwidth}{!}{
\begin{tabular}{l cccccccccc}
\toprule
 & \multicolumn{2}{c}{MNIST} & \multicolumn{2}{c}{Fashion-MNIST} & \multicolumn{2}{c}{CIFAR-10} & \multicolumn{2}{c}{TissueMNIST} & \multicolumn{2}{c}{PathMNIST} \\
\cmidrule(lr){2-3} \cmidrule(lr){4-5} \cmidrule(lr){6-7} \cmidrule(lr){8-9} \cmidrule(lr){10-11}
Algorithm & ARI & Accuracy & ARI & Accuracy & ARI & Accuracy & ARI & Accuracy & ARI & Accuracy \\
\midrule
Oracle   & $1.00 \pm 0.00$ & $96.45 \pm 0.38$ & $1.00 \pm 0.00$ & $81.90 \pm 0.82$ & $1.00 \pm 0.00$ & $56.52 \pm 1.57$ & $1.00 \pm 0.00$ & $27.37 \pm 6.12$ & $1.00 \pm 0.00$ & $45.57 \pm 8.49$ \\
\midrule
FedAvg   & --- & $83.65 \pm 1.81$ & --- & $63.41 \pm 3.26$ & --- & $\underline{45.64 \pm 6.39}$ & --- & $13.06 \pm 6.01$ & --- & $29.99 \pm 3.31$ \\
\midrule
FedGroup (s)& $0.02 \pm 0.07$ & $76.35 \pm 5.87$ & $0.19 \pm 0.26$ & $65.94 \pm 10.13$ & $-0.02 \pm 0.01$ & $42.93 \pm 2.77$ & $0.40 \pm 0.22$ & $25.02 \pm 3.46$ & $0.14 \pm 0.07$ & $38.50 \pm 5.39$ \\
StoCFL (s)& $-0.02 \pm 0.00$ & $64.67 \pm 1.88$ & $-0.01 \pm 0.01$ & $50.05 \pm 1.17$ & $-0.03 \pm 0.00$ & $31.04 \pm 4.12$ & $-0.02 \pm 0.00$ & $14.43 \pm 3.10$ & $-0.03 \pm 0.01$ & $26.37 \pm 2.81$ \\
FeSEM (s)& $0.04 \pm 0.08$ & $80.13 \pm 4.57$ & $0.20 \pm 0.13$ & $65.83 \pm 6.50$ & $-0.00 \pm 0.01$ & $42.84 \pm 1.09$ & $0.26 \pm 0.06$ & $20.50 \pm 0.81$ & $0.14 \pm 0.14$ & $33.74 \pm 5.34$ \\
IFCA (c)& $\mathbf{1.00 \pm 0.00}$ & $\underline{93.62 \pm 0.29}$ & $\mathbf{1.00 \pm 0.00}$ & $\underline{77.07 \pm 0.42}$ & $0.36 \pm 0.10$ & $42.21 \pm 5.33$ & $0.23 \pm 0.00$ & $16.75 \pm 0.00$ & $0.09 \pm 0.08$ & $27.54 \pm 2.63$ \\
K-Fed (m)& $\mathbf{1.00 \pm 0.00}$ & $88.56 \pm 1.09$ & $\mathbf{1.00 \pm 0.00}$ & $71.44 \pm 1.76$ & $\mathbf{1.00 \pm 0.00}$ & $35.90 \pm 2.40$ & $\mathbf{1.00 \pm 0.00}$ & $\underline{26.77 \pm 4.28}$ & $\mathbf{1.00 \pm 0.00}$ & $\underline{42.76 \pm 3.65}$ \\
PACFL (m)& $\mathbf{1.00 \pm 0.00}$ & $88.86 \pm 0.56$ & $\mathbf{1.00 \pm 0.00}$ & $72.38 \pm 1.30$ & $0.00 \pm 0.00$ & $33.53 \pm 8.73$ & $0.48 \pm 0.00$ & $10.59 \pm 2.37$ & $0.35 \pm 0.18$ & $37.20 \pm 7.21$ \\
FLAMECHE (m)& $\underline{0.96 \pm 0.04}$ & $\mathbf{94.95 \pm 2.07}$ & $\underline{0.99 \pm 0.03}$ & $\mathbf{81.02 \pm 1.52}$ & $\underline{0.97 \pm 0.03}$ & $\mathbf{57.07 \pm 3.13}$ & $\mathbf{1.00 \pm 0.00}$ & $\mathbf{29.56 \pm 0.00}$ & $\underline{0.96 \pm 0.05}$ & $\mathbf{44.11 \pm 3.11}$ \\
\bottomrule
\end{tabular}
}
\end{table}
In contrast, several baselines exhibit high sensitivity to the type of distribution shift. Methods such as PACFL or StoCFL perform well under isolated setups but degrade significantly on more complex datasets with combined heterogeneity. For example, under CIFAR-10 with concept shift on features combined with label skew, algorithms become confused and cluster clients based on label distribution rather than concept shift, which significantly impacts performance. Similarly, K-Fed achieves strong clustering quality once clustering is performed, but its one-shot strategy delays cluster formation, as it requires collecting metadata from all clients before clustering. Because it performs standard FedAvg prior to clustering, this suboptimal training dynamic negatively affects model learning in subsequent rounds.

FLAMECHE, by contrast, identifies cluster structure early through its EM formulation and continuously refines clusters during training. It starts clustering as soon as clients enter the federation, relying on rich low-dimensional metadata, which facilitates early structure discovery. Clustering in this space relies on simple distance computations, making it computationally efficient and avoiding the challenges of high-dimensional representations. This leads to stable performance across all settings and improved robustness to heterogeneous data distributions.

These observations are consistent with the aggregated results reported in the main paper (Table~\ref{tab:aggregated_performance}) and the aligned rank analysis (Figure~\ref{fig:rank}), where FLAMECHE achieves the best overall ranking across datasets and heterogeneity types.

\section{Hyperparameter Selection via Unsupervised Metrics}\label{sec:tuning}
Hyperparameter tuning is costly in federated settings~\cite{khodak2021federated}. The introduction of clustering in CFL further increases the number of hyperparameters, making this process even more costly. Selecting the number of clusters $K$ (or threshold-based hyperparameters that implicitly determine this number) in CFL typically requires running multiple end-to-end federated training procedures. Under most CFL algorithms (server-side and client-side), each configuration can only be evaluated using downstream metrics (e.g., test accuracy), leading to substantial communication and computation overhead dedicated solely to hyperparameter tuning. Since FLAMECHE decouples clustering from model training, hyperparameters can be evaluated directly on static metadata before any optimization, providing a satisfying heuristic.

We use the Davies-Bouldin Index (DBI) as an unsupervised selection criterion. DBI measures the ratio of intra-cluster dispersion to inter-cluster separation and can be computed from metadata alone while respecting privacy requirements. 

\begin{figure}[htb!]
    \centering
    \includegraphics[width=\textwidth]{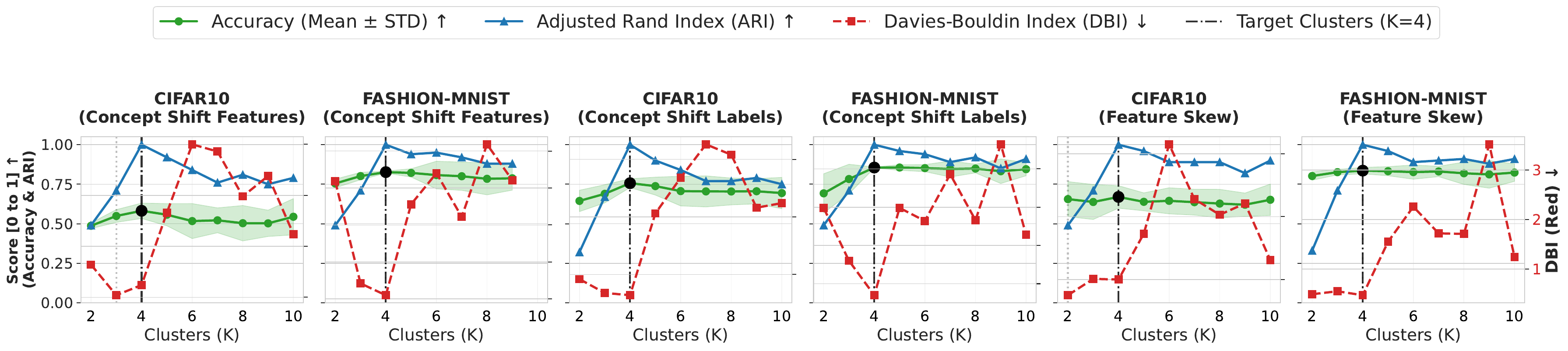}
    \caption{Hyperparameter selection before FL training. The Davies-Bouldin Index (DBI) is evaluated over different $K$. In most cases, low DBI aligns with the best  ARI and accuracy.}
    \label{fig:decouple}
\end{figure}
As shown in Figure~\ref{fig:decouple}, lower DBI values generally correspond to higher-quality clusterings, with good alignment to the best ARI and accuracy in most settings. While the configuration with the minimum DBI is not always the optimal one (e.g., under CIFAR10 with feature-distribution skew), DBI consistently narrows the search space to a small set of competitive candidates. This should be viewed as an empirical observation consistent across our experimental setups rather than a general guarantee.  This enables efficient hyperparameter selection without accessing labels or performing additional training rounds.

Crucially, DBI can be evaluated without exposing plaintext metadata. Each client computes its intra-cluster and inter-cluster distances locally and encrypts these quantities before transmission. The server then performs only additive aggregations over encrypted values, while non-linear operations (e.g., divisions and ratios) are deferred to the client side. This decomposition ensures full compatibility with privacy-preserving constraints.
\section{Robustness Analysis}
\subsection{Handling Missing Labels}\label{app:missing-label}

FLAMECHE relies on metadata that capture class-wise feature statistics, corresponding to the relationship between features and labels ($P(X|Y)$). As defined in Equation~\ref{eq:metadata}, each client represents its data through class-wise empirical means in the latent space of the feature extractor.

In practice, some clients may not observe all labels. To handle this, missing entries in the metadata vector are accounted for during the E-step by computing responsibilities only over the indices corresponding to observed classes. This ensures that the assignment step remains well-defined despite incomplete metadata.

To maintain a consistent representation dimensionality across clients, missing entries are then imputed using the nearest cluster centroid. Concretely, for a client representation $\phi_i = [\mu_{i,1}, \dots, \mu_{i,C}]$, if a class $c^*$ is missing, the corresponding feature $\mu_{i,c^*}$ is replaced by the value from the assigned centroid $\theta_{k_i}$ at round $r$. Since centroids are computed as the average of clients' metadata within a cluster at round $r-1$, this procedure amounts to replacing missing features with those of similar clients.

\begin{figure}[H]
    \centering
    \includegraphics[width=0.8\textwidth]{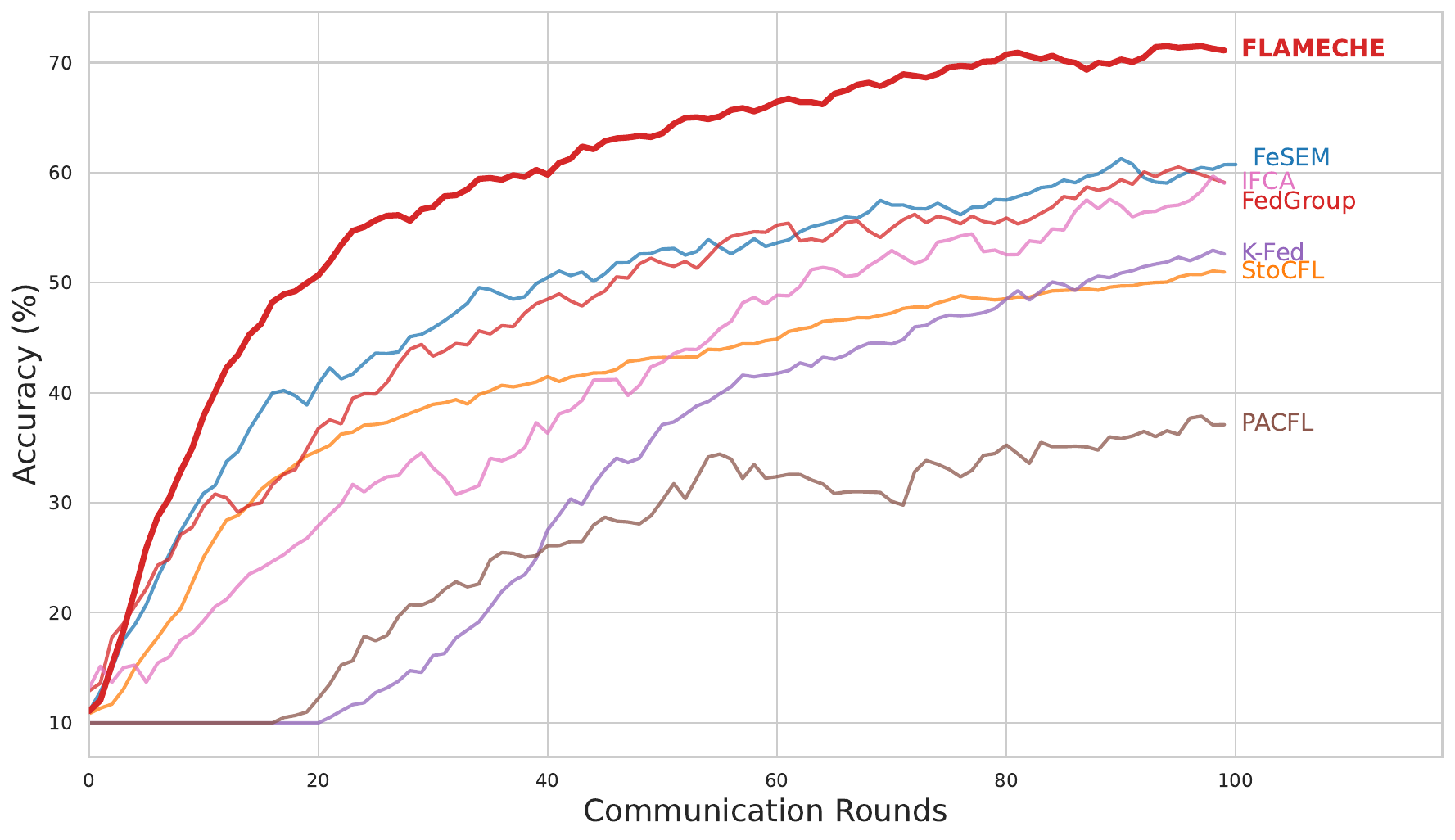}
        \caption{Accuracy per round under concept shift on features with missing labels under CIFAR10}
    \label{fig:miss_rounds}
\end{figure}

Empirically, this strategy preserves clustering quality even in the presence of missing labels. In Table~\ref{tab:concept_shift_missing_labels}, we evaluate a setting with concept shift on features where each client is missing one label. Results show that FLAMECHE remains competitive and is able to recover the correct clustering structure despite incomplete local label support. Figure~\ref{fig:miss_rounds} further illustrates that, although K-Fed can identify the true clusters, repeated training rounds with FedAvg degrade model performance in this setting.

\begin{table}[htb!]
\centering
\caption{Illustration of robustness under concept shift on features with one missing label per client. Performance comparison of FLAMECHE against baselines on Fashion-MNIST and CIFAR-10 datasets.}
\label{tab:concept_shift_missing_labels}
\begin{tabular}{l cccc}
\toprule
& \multicolumn{2}{c}{Fashion-MNIST} & \multicolumn{2}{c}{CIFAR-10} \\
\cmidrule(lr){2-3} \cmidrule(lr){4-5}
Method & ARI & Accuracy (\%) & ARI & Accuracy (\%) \\
 \midrule
StoCFL   & $\mathbf{1.00}$ & $81.96 \pm 0.65$ & $0.00$ & $52.33 \pm 8.61$ \\
FedGroup & $\mathbf{1.00}$ & $\mathbf{86.34 \pm 0.59}$ & $0.07$ & $57.01 \pm 7.30$ \\
FeSEM    & $\mathbf{1.00}$ & $86.11 \pm 0.61$ & $0.18$ & $\underline{60.01 \pm 6.77}$ \\
IFCA     & $\mathbf{1.00}$ & $81.21 \pm 3.91$ & $\underline{0.71}$ & $58.88 \pm 4.30$ \\
PACFL    & $\mathbf{1.00}$ & $76.86 \pm 0.57$ & $0.07$ & $36.26 \pm 5.72$ \\
K-fed & $\mathbf{1.00}$ & $75.61 \pm 0.47$ & $\mathbf{1.00}$ & $50.00 \pm 9.28$ \\

FLAMECHE & $\mathbf{1.00}$ & $\underline{86.33 \pm 1.01}$ & $\mathbf{1.00}$ & $\mathbf{69.90 \pm 2.93}$ \\
\bottomrule
\end{tabular}
\end{table}

\subsection{Reclustering}\label{app:recluster}

FLAMECHE relies on a distributed EM procedure over client metadata. In standard EM algorithms, empty cluster configurations are a known issue~\cite{zhang2003algorithms}, i.e., clusters that receive no assignments during the E-step. While transient empty clusters may occur without affecting the procedure, clusters that remain empty across iterations lead to degenerate solutions. This is particularly critical in our setting, as FLAMECHE cannot leverage methods that require direct observation of metadata~\cite{arthur2007k,zhang2003algorithms}; metadata representations are preferably not directly accessible.

To mitigate this limitation, we introduce a \textit{reclustering mechanism} (in Algorithm \ref{alg:flameche}) that periodically reinitializes empty clusters. Specifically, every $\tau$ communication rounds, empty clusters and clusters with maximum average error are reinitialized. This allows the algorithm to prevent clusters from remaining inactive throughout training.

\begin{table}[ht]
\centering
\resizebox{\linewidth}{!}{
\begin{tabular}{llcccccc}
\toprule
\multirow{2}{*}{Heterogeneity} & \multirow{2}{*}{Reclustering} & \multicolumn{2}{c}{Fashion-MNIST} & \multicolumn{2}{c}{PathMNIST} & \multicolumn{2}{c}{CIFAR-10} \\
\cmidrule(lr){3-4} \cmidrule(lr){5-6} \cmidrule(lr){7-8}
 & & ARI & DBI & ARI & DBI & ARI & DBI \\
\midrule

Concept Shift & $\tau = 10 $& $0.97 \pm 0.05$ & $0.50 \pm 0.02$ & $0.97 \pm 0.05$ & $0.92 \pm 0.25$ & $0.96 \pm 0.04$ & $1.53 \pm 0.16$ \\
on Features & $\tau = 20 $& $0.97 \pm 0.05$ & $0.47 \pm 0.04$ & $0.96 \pm 0.04$ & $0.92 \pm 0.26$ & $0.97 \pm 0.05$ & $1.53 \pm 0.14$ \\
+ Label Skew & $\tau = 30 $& $1.00 \pm 0.00$ & $0.43 \pm 0.04$ & $1.00 \pm 0.00$ & $0.79 \pm 0.10$ & $1.00 \pm 0.00$ & $1.45 \pm 0.17$ \\
& No & $0.89 \pm 0.18$ & $0.70 \pm 0.51$ & $0.80 \pm 0.18$ & $0.72 \pm 0.12$ & $1.00 \pm 0.00$ & $1.50 \pm 0.18$ \\
\midrule

Concept Shift
 & $\tau = 10 $ & $0.97 \pm 0.05$ & $0.41 \pm 0.07$ & $0.97 \pm 0.05$ & $1.51 \pm 0.13$ & $0.97 \pm 0.05$ & $1.95 \pm 0.11$ \\
 on Labels & $\tau = 20 $ & $0.97 \pm 0.05$ & $0.39 \pm 0.09$ & $0.97 \pm 0.05$ & $1.69 \pm 0.28$ & $0.97 \pm 0.05$ & $1.93 \pm 0.15$ \\
 & $\tau = 30 $ & $1.00 \pm 0.00$ & $0.33 \pm 0.08$ & $1.00 \pm 0.00$ & $1.67 \pm 0.28$ & $1.00 \pm 0.00$ & $1.87 \pm 0.13$ \\
& No & $1.00 \pm 0.00$ & $0.33 \pm 0.08$ & $1.00 \pm 0.00$ & $1.67 \pm 0.28$ & $0.88 \pm 0.21$ & $2.20 \pm 0.56$ \\
\midrule

Features
 & $\tau = 10 $ & $0.97 \pm 0.05$ & $0.51 \pm 0.05$ & $0.97 \pm 0.05$ & $0.68 \pm 0.11$ & $0.98 \pm 0.04$ & $1.08 \pm 0.05$ \\
 Distribution & $\tau = 20 $ & $0.97 \pm 0.05$ & $0.52 \pm 0.03$ & $0.97 \pm 0.05$ & $0.76 \pm 0.11$ & $0.97 \pm 0.05$ & $1.07 \pm 0.11$ \\
 Skew & $\tau = 30 $ & $1.00 \pm 0.00$ & $0.49 \pm 0.03$ & $1.00 \pm 0.00$ & $0.73 \pm 0.10$ & $1.00 \pm 0.00$ & $1.03 \pm 0.06$ \\
  & No & $0.87 \pm 0.22$ & $1.02 \pm 0.92$ & $1.00 \pm 0.00$ & $0.73 \pm 0.10$ & $0.89 \pm 0.19$ & $1.41 \pm 0.69$ \\
\bottomrule
\end{tabular}
}
\caption{Reclustering comparison for different heterogeneity classes and parameters tested over 3 random seeds. Reclustering happens every $\tau$ rounds when having empty clusters. 
}
\label{tab:combined_performance}
\end{table}

Table~\ref{tab:combined_performance} reports the impact of different reclustering frequencies ($\tau \in \{10,20,30\}$) as well as the case without reclustering. Results are averaged over $3$ random seeds across different heterogeneity settings.

Overall, reclustering consistently improves clustering quality compared to no reclustering, especially under more complex heterogeneity, such as combined concept shift on features with label skew and feature distribution skew. Without reclustering, dead clusters persist in several runs, leading to degraded ARI and higher DBI (i.e., poorer cluster compactness). 

Across all settings, moderate reclustering frequencies ($\tau = 20$ or $\tau = 30$) provide the most stable results. Smaller values (e.g., $\tau=10$) already mitigate dead clusters but may introduce slight instability due to more frequent reinitializations. In contrast, larger values cause reclustering to happen late in the federation, with risks of negative impact on learning.

These results highlight that such reclustering is a simple yet effective mechanism to improve the robustness of FLAMECHE.

\subsection{Metadata Accumulation and Partial Participation.}\label{app:partial}
In FLAMECHE, we recommend computing cluster centroids during the M-step over the full set of \emph{seen} clients, with their metadata stored at the server. Because client metadata representations remain static across communication rounds, the server can accumulate and store these representations over time, even in encrypted form. As new clients participate, their metadata and cluster assignments are incorporated, allowing the clustering structure to refine progressively. At each communication round, only the assignments of participating clients are updated.

However, accumulating metadata across rounds may introduce complexities under certain cryptographic protocols. For instance, in Secure Aggregation~\cite{bonawitz2017practical}, masks are designed to cancel within a single round, making persistent cross-round statistics less straightforward to maintain without additional mechanisms. As a result, this setting does not constitute a direct application of standard Secure Aggregation workflows.

To ensure easier compatibility with such protocols, FLAMECHE can be restricted to computing the M-step using only the metadata of clients participating in the current round. Table~\ref{tab:flameche_partial_vs_normal} provides an illustrative comparison between these two strategies. While using only current-round participants may lead to slightly degraded clustering quality in some settings, leveraging the full set of seen clients generally provides more stable results. Overall, the differences remain limited in our experiments, indicating that the partial variant remains a viable alternative when required by the deployment setting. This observation, however, may not hold under very low participation rates, where limited client coverage per round can affect the stability of the estimated cluster statistics.

\begin{table}[htb!]
\centering
\caption{Example of comparison of FLAMECHE clustering performance (ARI and DBI) using all seen metadata (normal) versus only round participants (partial) for the maximization step. Results are shown for an independent run that highlight the difference between the two approaches.}
\label{tab:flameche_partial_vs_normal}
\resizebox{\textwidth}{!}{
\begin{tabular}{ll cc cc cc cc cc}
\toprule
& & \multicolumn{2}{c}{\textbf{MNIST}} & \multicolumn{2}{c}{\textbf{Fashion-MNIST}} & \multicolumn{2}{c}{\textbf{TissueMNIST}} & \multicolumn{2}{c}{\textbf{PathMNIST}} & \multicolumn{2}{c}{\textbf{CIFAR-10}} \\
\cmidrule(lr){3-4} \cmidrule(lr){5-6} \cmidrule(lr){7-8} \cmidrule(lr){9-10} \cmidrule(lr){11-12}
\textbf{Algorithm} & \textbf{Metadata Aggregation} & \textbf{ARI} & \textbf{DBI} & \textbf{ARI} & \textbf{DBI} & \textbf{ARI} & \textbf{DBI} & \textbf{ARI} & \textbf{DBI} & \textbf{ARI} & \textbf{DBI} \\
\midrule
\multirow{2}{*}{\textbf{FLAMECHE}} & \textbf{Normal (All seen)} & 1.00 & 0.58 & 1.00 & 0.42 & 1.00 & 1.15 & 1.00 & 0.97 & 1.00 & 1.53 \\
& \textbf{Partial (Participants)} & 1.00 & 0.58 & 1.00 & 0.42 & 1.00 & 1.15 & 0.98 & 1.03 & 0.97 & 1.67 \\
\bottomrule
\end{tabular}
}
\end{table}

\section{FLAMECHE under Different Cryptographic Settings}\label{app:crypto}

While FLAMECHE is agnostic to specific additive cryptographic schemes, it is essential to discuss the potential implications of each deployment context. In this section, we consider three of the most practical secure FL schemes~\cite{mothukuri2021survey}: Secure Aggregation~\cite{bonawitz2017practical}, Paillier Homomorphic Encryption~\cite{wang2024priverifl}, and CKKS Homomorphic Encryption~\cite{cheon2017homomorphic}. In this Appendix section, we focus on enabling encryption for metadata only; encryption of model updates follows the standard FedAvg setting and is therefore not discussed further.

\subsection{Secure Aggregation}\label{app:sa}
In standard Secure Aggregation protocols~\cite{bonawitz2017practical}, masking terms cancel out exactly during aggregation, allowing the server to directly recover the sum of the metadata in plaintext. Consequently, the server can update the centroids in Equation~\ref{eq:M-step} without relying on clients. Masks are constructed to cancel pairwise between clients, and in practice, this cancellation is ensured within each round among participating clients. As clients join and leave the federation, the protocol naturally guarantees correct cancellation for current-round participants.

In the CFL setting, this implies that masking must be established accordingly: at a given round, clients should form masks only with other participants assigned to the same cluster. As mentioned in Appendix~\ref{app:partial}, under FLAMECHE with Secure Aggregation, this is achieved by restricting the M-step to current-round participants. In this case, since the original Secure Aggregation protocol already accounts for dynamic participation, applying it in the CFL setting amounts to considering each cluster as a separate FL instance, where clients join and leave over time.

We emphasize that this requirement concerns the more complex, potentially persistent masking management of the Secure Aggregation protocol under M-step over all seen clients, rather than a flaw in the FLAMECHE algorithm.

\subsection{Homomorphic Encryption}\label{app:he}

Under Homomorphic Encryption (HE), the server operates entirely on encrypted values and cannot decrypt aggregated results. In this case, the server have to rely on clients to update the centroids in Equation~\ref{eq:M-step}. For each cluster, because the aggregate $[[S_k]]$ remains encrypted, the server needs to broadcast $[[S_k]]$ and $|C_k^{(r)}|$ instead of the centroids. Clients can decrypt and compute Equation~\ref{eq:M-step} locally to proceed to E-step. Because the updated centroids are only utilized by the clients during the E-step, this delegation does not disrupt the clustering workflow.

\paragraph{Paillier Homomorphic Encryption.}
Paillier is an additive homomorphic encryption scheme defined over integers. In practice, real-valued quantities (e.g., gradients or metadata) must therefore be encoded into integers via fixed-point scaling before encryption. While the management of key initialization is a well-studied problem in the literature~\cite{mothukuri2021survey}. The computational cost of Paillier is non-trivial~\cite{wang2024priverifl}, especially if used over extreme dimension model updates. While protocols~\cite{wang2024priverifl, wang2024fvfl} exist to take account of those high dimensions, as shown in our timing analysis (~\ref{app:crypt_disc}), in FLAMEHCE the latency overhead is manageable exclusively because it operates on low-dimensional metadata rather than full model weights. Paillier results use fixed-point encoding ($10^9$ scale), hence represent approximate real-valued aggregation. Empirically, this quantization has no observable impact on FLAMECHE performance across all evaluated datasets, yielding results identical to the plaintext baseline. 

\paragraph{CKKS Homomorphic Encryption.}
CKKS~\cite{cheon2017homomorphic,pan2024fedshe} is an efficient partial HE scheme designed for floating-point arithmetic, which introduces a small numerical perturbation (noise) during encoding, rescaling, and aggregation. Table~\ref{tab:ckks_ablation} highlights the empirical impact of this CKKS noise on FLAMECHE's clustering performance. 

For this evaluation, CKKS is implemented using the TenSEAL Python library with standard parameters ($N=16384,\ \{60,50,50,60\},\ \Delta=2^{50}$), while Paillier uses the \texttt{phe} Python library. The numerical perturbation affecting client assignments is negligible in most datasets (typically $\le 0.02$ ARI difference). However, the perturbation has a slightly more pronounced effect on the TissueMNIST dataset. This occurs because the decision boundaries between heterogeneous groups in this specific setting are more subtle and thus more sensitive to cryptographic noise. Refining the reclustering mechanism or increasing the precision of the CKKS parameters could further mitigate these effects.

\begin{table}[htb!]
\centering
\caption{Impact of CKKS numerical noise on clustering performance of FLAMECHE. Results compare the unencrypted baseline (no-CKKS) against execution under Partial Homomorphic Encryption (CKKS).}
\label{tab:ckks_ablation}
\resizebox{\textwidth}{!}{
\begin{tabular}{l cccccccccc}
\toprule
& \multicolumn{2}{c}{MNIST} & \multicolumn{2}{c}{Fashion-MNIST} & \multicolumn{2}{c}{CIFAR-10} & \multicolumn{2}{c}{TissueMNIST} & \multicolumn{2}{c}{PathMNIST} \\
\cmidrule(lr){2-3} \cmidrule(lr){4-5} \cmidrule(lr){6-7} \cmidrule(lr){8-9} \cmidrule(lr){10-11}
Setting & ARI & DBI & ARI & DBI & ARI & DBI & ARI & DBI & ARI & DBI \\
\midrule
no-CKKS & $0.96 \pm 0.04$ & $0.60 \pm 0.09$ & $0.97 \pm 0.05$ & $0.43 \pm 0.09$ & $0.97 \pm 0.05$ & $1.49 \pm 0.37$ & $1.00 \pm 0.00$ & $1.09 \pm 0.52$ & $0.95 \pm 0.11$ & $1.02 \pm 0.58$ \\
CKKS    & $0.94 \pm 0.13$ & $0.61 \pm 0.12$ & $0.96 \pm 0.11$ & $0.46 \pm 0.15$ & $0.96 \pm 0.11$ & $1.59 \pm 0.36$ & $0.85 \pm 0.17$ & $1.35 \pm 0.46$ & $0.93 \pm 0.14$ & $1.07 \pm 0.53$ \\
\bottomrule
\end{tabular}
}
\end{table}

\subsection{Discussion of Cryptographic Implications}\label{app:crypt_disc}
\textit{This section provides a high-level discussion of cryptographic implications. The reported costs are intended to give general insights and do not account for implementation-specific optimizations or protocol-level communication details, which are outside the scope of this work.}

\begin{table}[htb!]
\centering
\caption{Per-round cryptographic overhead for 100 clients on a standard workstation (Intel Core i7, 16 threads). Results are reported for full-model updates and metadata-based aggregation. Secure Aggregation reports masking and aggregation costs. CKKS is implemented using TenSEAL, encrypting each vector as a single ciphertext. Paillier results are obtained using our optimized C++ implementation (2048-bit keys, GMP backend). ResNet-18 metadata timings are measured; LeNet-5 metadata timings are linearly scaled from these measurements ($\dagger$). Implementations follow standard libraries for each scheme and are therefore representative rather than strictly implementation-matched.}
\label{tab:crypto_benchmarks}
\resizebox{\textwidth}{!}{
\begin{tabular}{ll cc | ccc | ccc}
\toprule
& & \multicolumn{2}{c|}{\textbf{Secure Aggregation}} & \multicolumn{3}{c|}{\textbf{HE (CKKS)}} & \multicolumn{3}{c}{\textbf{HE (Paillier)}} \\
\cmidrule(lr){3-4} \cmidrule(lr){5-7} \cmidrule(lr){8-10}
\textbf{Architecture} & \textbf{Representation} & \textbf{Client (Mask)} & \textbf{Server (Sum)} & \textbf{Client (Enc)} & \textbf{Server (Sum)} & \textbf{Client (Dec)} & \textbf{Client (Enc)} & \textbf{Server (Sum)} & \textbf{Client (Dec)} \\
\midrule
\textbf{LeNet-5} 
& Full Model & 27.72 ms & 2.80 ms & 60.70 ms & 99.20 ms & 14.90 ms & -- & -- & -- \\
& \textbf{Metadata} & \textbf{0.35 ms} & \textbf{0.05 ms} & \textbf{6.90 ms} & \textbf{12.40 ms} & \textbf{1.80 ms} & \textbf{9.57 s}$^{\dagger}$ & \textbf{0.33 s}$^{\dagger}$ & \textbf{6.22 s}$^{\dagger}$ \\
\midrule
\textbf{ResNet-18} 
& Full Model & 5.39 s & 0.93 s & 10.33 s & 16.93 s & 2.60 s & -- & -- & -- \\
& \textbf{Metadata} & \textbf{2.54 ms} & \textbf{0.26 ms} & \textbf{9.00 ms} & \textbf{12.40 ms} & \textbf{1.80 ms} & \textbf{58.30 s} & \textbf{2.01 s} & \textbf{37.89 s} \\
\bottomrule
\end{tabular}
}
\end{table}

\begin{table}[htb!]
\centering
\caption{On-client communication cost for metadata and full-model representations. Secure Aggregation assumes 64-bit masks (8 bytes per parameter). CKKS is evaluated using TenSEAL ($N=16384$, scale $2^{50}$), encrypting each vector as a single ciphertext. Paillier uses our C++ implementation with per-element encryption; ciphertext sizes are measured via hex encoding. Reported sizes exclude protocol-specific overhead.}
\label{tab:communication_cost}
\resizebox{0.85\textwidth}{!}{
\begin{tabular}{@{}llrrrrr@{}}
\toprule
\textbf{Architecture} & \textbf{Representation} & \textbf{Dimension} & \textbf{Plaintext Size} & \textbf{Secure Aggregation (mask)} & \textbf{HE (CKKS)} & \textbf{HE (Paillier)} \\ 
\midrule
\textbf{LeNet-5} 
& Full Model & 61,706 & 0.2404 MB & -- & -- & -- \\
& \textbf{Metadata} & \textbf{840} & \textbf{0.0032 MB} & \textbf{0.0064 MB} & \textbf{0.6996 MB} & \textbf{0.4095 MB} \\ 
\midrule
\textbf{ResNet-18} 
& Full Model & 11,176,512 & 42.7280 MB & -- & -- & -- \\
& \textbf{Metadata} & \textbf{5,120} & \textbf{0.0195 MB} & \textbf{0.0391 MB} & \textbf{0.6996 MB} & \textbf{2.4963 MB} \\ 
\bottomrule
\end{tabular}
}
\end{table}

Table~\ref{tab:crypto_benchmarks} reports the wall-clock time for applying Secure Aggregation, Paillier, and CKKS HE protocols, evaluated on a standard workstation (Intel Core i7, 16 threads). Encrypting full model weights introduces substantial latency. 

For Paillier, this cost is further amplified by the lack of vector encryption and the need for per-element encoding; as a result, full-model encryption with Paillier is prohibitively expensive and is therefore not reported. In our implementation, encryption and decryption of metadata are not cheap, but depending on the setup, this may appear acceptable compared to local model training. Methods exist to lower these costs~\cite{wang2024priverifl, wang2024fvfl,mothukuri2021survey}, but their study and impact on FLAMECHE are out of the scope of this paper. 

For CKKS HE specifically, large models exceed single-CKKS ciphertext capacity (e.g., $8,192$ slots in TenSEAL CKKS), requiring severe fragmentation that drives server-side summation times up to $17$ seconds for ResNet-18. In contrast, FLAMECHE metadata fits comfortably within a single ciphertext. This results in millisecond-scale overhead for encryption, masking, and aggregation. As detailed in Table~\ref{tab:communication_cost}, the additional payload of encrypted metadata becomes negligible compared to multiple model sharings as models become increasingly larger. This dimensionality reduction ensures that strictly encrypted clustering remains highly practical without requiring modifications to standard cryptographic pipelines.

Ultimately, these benchmarks highlight how FLAMECHE effectively navigates the CFL trilemma. Historically, deploying HE or SMPC for server-side clustering was considered prohibitively expensive, forcing practitioners to either abandon strong privacy guarantees (by sharing plaintext metadata) or accept severe computational bottlenecks (by relying on encrypted clustering). By compressing the clustering signal into low-dimensional metadata and strictly bounding server-side operations to linear aggregations, FLAMECHE bridges this gap. It demonstrates that metadata-based clustering can be deployed efficiently under standard privacy-preserving FL protocols.

\end{document}